\pdfoutput=1
\documentclass[11pt]{article}

\usepackage{acl}

\usepackage{scrextend}
\usepackage{colortbl}

\usepackage{times}
\usepackage{latexsym}
\usepackage{graphicx}
\usepackage{booktabs}
\usepackage{multirow}
\usepackage{arydshln}
\usepackage{comment}
\usepackage{multirow}
\usepackage{amssymb}
\usepackage{amsmath}
\usepackage{bclogo}
\usepackage[skins]{tcolorbox}
\usepackage{rotating}
\usepackage{graphicx}
\usepackage[T1]{fontenc}
\usepackage{footmisc}
\usepackage{todonotes}

\usepackage[utf8]{inputenc}

\usepackage{microtype}

\usepackage{inconsolata}

\usepackage{graphicx}

\newcolumntype{L}[1]{>{\raggedright\let\newline\\\arraybackslash\hspace{0pt}}m{#1}}
\newcolumntype{C}[1]{>{\centering\let\newline\\\arraybackslash\hspace{0pt}}m{#1}}
\newcolumntype{R}[1]{>{\raggedleft\let\newline\\\arraybackslash\hspace{0pt}}m{#1}}

\newcommand\Lou{\texttt{Lou}}
\title{The \Lou{} Dataset \\ Exploring the Impact of Gender-Fair Language in German Text Classification }

\author{Andreas Waldis\thanks{The corresponding author is andreas.waldis@live.com} $^{1,2}$, 
Joel Birrer$^{2}$,
Anne Lauscher$^{3}$,
Iryna Gurevych$^{1}$\\
$^1$Ubiquitous Knowledge Processing Lab (UKP Lab) \\
Department of Computer Science and Hessian Center for AI (hessian.AI)\\
Technical University of Darmstadt\\
 $^2$Information Systems Research Lab, Lucerne University of Applied Sciences and Arts \\
 $^3$Data Science Group, University of Hamburg\\
\texttt{\href{http://www.ukp.tu-darmstadt.de/}{www.ukp.tu-darmstadt.de}} \hspace{0.5em} \texttt{\href{http://www.hslu.ch/}{www.hslu.ch}}\\
}
\definecolor{lightgray}{gray}{0.9}

\newcommand\Doppelnennung{\texttt{Doppelnennung}}
\newcommand\GenderStern{\texttt{GenderStern}}
\newcommand\GenderGap{\texttt{GenderGap}}
\newcommand\GenderDoppelpunkt{\texttt{GenderDoppelpunkt}}
\newcommand\Neutral{\texttt{Neutral}}
\newcommand\Dee{\texttt{De-e}}
\begin{document}
\maketitle
\vspace*{1em}
\begin{abstract}
Gender-fair language, an evolving German linguistic variation, fosters inclusion by addressing all genders or using neutral forms. 
Nevertheless, there is a significant lack of resources to assess the impact of this linguistic shift on classification using language models (LMs), which are probably not trained on such variations.
To address this gap, we present \Lou{}, the first dataset featuring high-quality reformulations for German text classification covering seven tasks, like stance detection and toxicity classification. 
Evaluating 16 mono- and multi-lingual LMs on \Lou{} shows that gender-fair language substantially impacts predictions by flipping labels, reducing certainty, and altering attention patterns. 
However, existing evaluations remain valid, as LM rankings of original and reformulated instances do not significantly differ.
While we offer initial insights on the effect on German text classification, the findings likely apply to other languages, as consistent patterns were observed in multi-lingual and English LMs.\footnote{Data is also available in a \href{https://tudatalib.ulb.tu-darmstadt.de/handle/tudatalib/4350}{online archive}.}

%Our study provides initial insights into the impact of gender-fair language on classification for German.
%However, these findings are likely transferable to other languages, as we found consistent patterns in multi-lingual and English LMs.%\footnote{We release the \Lou{} dataset and code at \href{https://github.com/UKPLab/lou-gender-fair-reformulations}{https://github.com/UKPLab/lou-gender-fair-reformulations}.}
%existing classification datasets. 
%text classification examples like stance detection. 
%following six gender-inclusive and gender-neutralization strategies.
%\Lou{} covers test examples for seven German classification tasks like stance detectis

\textcolor{red}{Warning: This paper contains offensive text.}
\end{abstract}
\vspace{0em}
\includegraphics[width=2em,height=2em]{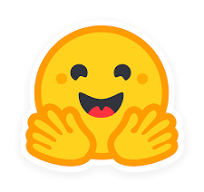}\hspace{.75em}\parbox{15em}{\href{https://huggingface.co/datasets/tresiwalde/lou}{\vspace*{1.3em}\small{\texttt{huggingface.co/datasets/tresiwalde/lou}}}
}
\newline
\includegraphics[width=2em,height=2em]{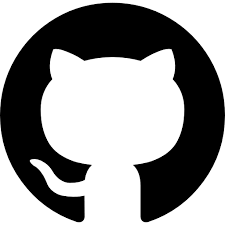}\hspace{.75em}\parbox{15em}{\href{https://github.com/UKPLab/lou-gender-fair-reformulations}{\vspace*{1.3em}\small{\texttt{UKPLab/lou-gender-fair-reformulations}}}
}
\vspace{-.5em}

\section{Introduction}\label{sec:introduction}

\begin{figure}[t]
\vspace*{2em}
    \centering
    \includegraphics[width=0.40\textwidth]{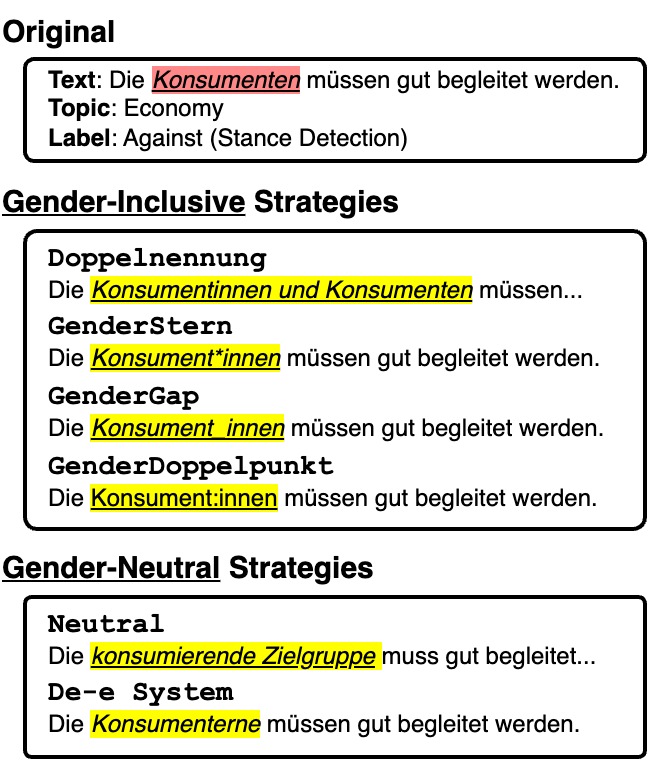}
    \caption{
    A German stance detection instance from the \Lou{} dataset.
    We reformulate the masculine formulation \textit{Konsumenten} (\textit{consumers}) regarding six inclusive or neutral strategies, highlighted in yellow.
    Translation: \textit{\underline{Consumers} must be well supported.}
    }
    \label{fig:overview}
\end{figure}

Language is constantly evolving.
This change involves dialect variations of specific localities or the slang of certain generations \citep{keidar-etal-2022-slangvolution, sun-xu-2022-tracing}.
Such linguistic variations reflect societal changes, where negotiations emerge and influence how individuals speak.
Notable shifts are gender-fair formulations in languages with feminine and masculine genders, like German or French.
The societal importance of such linguistic variations is reflected in concrete guidelines, like those from the EU Parliament.\footnote{Available \href{https://www.europarl.europa.eu/cmsdata/151780/GNL_Guidelines_EN.pdf}{online}.} 
Concretely, inclusive and neutral strategies \citep{Lardelli_Gromann}, illustrated in \autoref{fig:overview} for German, serve as tools to reduce gender stereotypes and discrimination \citep{10.3389/fpsyg.2016.00025} and to meet the UN goal of gender equality.\footnote{\href{https://web.archive.org/web/20240923052010/https://sdgs.un.org/goals/goal5}{https://sdgs.un.org/goals/goal5}.}
Inclusive strategies (\Doppelnennung{}) address both genders explicitly (\textit{Konsumentinnen und Konsumenten}, \textit{consumer.FEM.PL\footnote{In English, we indicate the German gender following the \href{https://web.archive.org/web/20240901180348/https://www.eva.mpg.de/lingua/pdf/Glossing-Rules.pdf}{Lepzig Glossing Rules}. We use \textit{FEM.MASC.NEUT} for formulations neutralizing or addressing any gender.} and consumer.MASC.PL}), or any gender using special characters (\GenderStern{}, \textit{Konsument*innen}, consumer.FEM.MASC.NEUT.PL).
Neutral strategies avoid specific genders using neutral terms (\textit{konsumierende Zielgruppe, consuming target group}). 

As language models (LMs) are inherently biased from training on data from the past \citep{kurita-etal-2019-measuring,DBLP:journals/corr/abs-2206-04615,attanasio-etal-2023-tale}, current research increasingly addresses the question `\textit{How does gender-fair language impact LMs?}`.
This includes examining the gender bias of machine translation regarding gender-fair language \citep{paolucci-etal-2023-gender,piergentili-etal-2023-gender} and pronouns \citep{lauscher-etal-2023-em,amrhein-etal-2023-exploiting}, or fundamentally concentrate on LMs' limitations when interpreting pronouns \citep{brandl-etal-2022-conservative,hossain-etal-2023-misgendered, Gautam2024RobustPU}. 
Despite the widespread application of text classification \citep{wang-etal-2018-glue, DBLP:conf/nips/ZhangZL15}, there is a notable lack of resources and scientific effort to examine the impact of gender-fair language on classification systems. 
Consequently, it remains unclear whether LMs perform consistently without unwanted side effects when processing gender-fair language under the inclusive aim.

To address this research gap, we introduce \Lou{} (\autoref{sec:dataset}), a German dataset featuring high-quality text reformulations following six reformulation strategies (\autoref{fig:overview}).
While creating these reformulations, we also examine the reliability of amateurs with moderate experience using gender-fair language compared to professionals with linguistic backgrounds. 
Using the resulting 3.6k reformulated instances from seven classification tasks, we systematically evaluate the impact of gender-fair language and specific reformulation strategies for fine-tuning and zero-shot classification setups covering 16 LMs (\autoref{sec:setup}).
We then compare these results with the original instances (\autoref{sec:results} and \autoref{sec:analysis}) to address four key research questions.

\paragraph{RQ1: Do amateurs produce gender-fair reformulations with sufficient quality?}
No. 
Amateurs struggle to consistently apply reformulation strategies, with an error rate of up to 31\% for \GenderStern{}, hinting at a lack of societal establishment and standardization.
%CR: highlight our categorization

\paragraph{RQ2: Does gender-fair language impact German text classification during inference?}
Yes.
Gender-fair language leads to task performance variations in macro $F_1$ score, ranging from -1.0 to +4.0, and flips up to 10.9\% of individual predictions.
However, the effects of distinct reformulation strategies vary. 
Those making minimal sentence adjustments, such as \GenderStern{}, tend to enhance performance. 
At the same time, neutralization-focused strategies, like \Dee{} or \Neutral{}, which substantially alter the text, generally lead to lower performance.

\paragraph{RQ3: Do LMs process gender-fair language differently?}
Yes. 
Gender-fair language notably impacts how the lower LM layers process reformulated instances compared to the original ones. 
Further, altered attention patterns and decreased prediction certainty lead to the observed label flips.

\paragraph{RQ4: What are the practical implications of encountering gender-fair language?}
Existing datasets and evaluations remain valid (consistent LM rankings), but significant label flips occur, especially in tasks with lower absolute performance. 
Observing mostly syntactic and consistent effects across German, English, and multi-lingual LMs, our findings are relevant to other languages with similar reformulation strategies, such as using the interpoint (·) as an inclusion character in French.

\paragraph{Contributions} 
We lay the foundation for studying the impact of gender-fair language on classification and make three key contributions:
%With a three-fold contribution, we lay the ground to study LMs under gender-fair language more comprehensively:
\textbf{1)} We present the first high-quality dataset of reformulated text instances for German classification tasks and provide insights into the practical annotation challenges.
\textbf{2)} A systematic evaluation underscores the practical value of German-specialized LMs and reveals the substantial impact of gender-fair language and individual reformulation strategies on individual predictions. 
\textbf{3)} We offer concrete guidance on how LMs process gender-fair language differently, highlighting the necessity to consider such fine-grained linguistic variations. 

\par

\section{Preliminaries}\label{sec:preliminaries}

\subsection{Gender-Fair Language}
We define \textit{gender-fair} language as a specific linguistic phenomenon that replaces the generic formulations, either the feminine or the predominant masculine one.
With alternative formulations, this linguistic shift reduces gender stereotypes and discrimination by comprehensively addressing people \citep{10.3389/fpsyg.2016.00025}.
As in \citet{Lardelli_Gromann}, gender-fair embodies both \textit{inclusive} and \textit{neutral} language (\autoref{fig:overview}).
Inclusive language addresses either the masculine or feminine gender explicitly or uses characters like the gender star \textbf{(*)} (German) to address everyone on the gender spectrum, including those identifying with no gender. 
Differently, neutral language prevents gender-specific formulations with alternative terms.

\subsection{Gender-Fair Reformulation Strategies}
Different strategies guide the formulation of gender-fair language.
Specifically, we consider the following inclusive or neutral ones.\footnote{As these strategies are proper names, we do not translate from German to English.}

\paragraph{i) Binary Gender Inclusion (\Doppelnennung{})} explicitly mentions the feminine and masculine but ignores others like agender. 
For example, \textit{Ärzte} (doctor.MASC.PL) is transformed into \textit{Ärztinnen und Ärzte} (\textit{doctor.FEM.PL and doctor.MASC.PL}).
%Note, this such formulation do not address other gendersSuch a formulation focuses on binary inclusion but not on diverse or gender. 

\paragraph{ii) All Gender Inclusion} explicitly addresses every gender, including agender, non-binary, or demi-gender, using a gender gap character pronounced with a small pause. %CR: footnote with link to an exhaustive list
%This character is pronounced with a small pause and accounts for the feminine, masculine, and every other gender on or outside  this spectrum. 
In this work, we consider three commonly used gender characters: \GenderStern{} (\textbf{*}), \GenderDoppelpunkt{} (\textbf{:}), and  \GenderGap{} (\textbf{\_}).
For example, \textit{Ärzte} (\textit{doctor.MASC.PL}) is turned into \textit{Ärzt*innen}, \textit{Ärzt:innen}, or \textit{Ärzt\_innen} (doctor.FEM.MASC.NEUT.PL).

\paragraph{iii) Gender Neutralization (\Neutral{})} avoids naming a particular gender using neutral terms, like \textit{ärztliche Fachperson} (\textit{medical professional}).% to prevent any gender. 
%For example, \textit{Arzt} (\textit{doctor.MASC.SG}) is turned into \textit{ärztliche Fachperson} (\textit{medical professional}).

\paragraph{iv) \Dee{} (Neosystem)} is a well-specified system that emerged from a significant community-driven effort.\footnote{Find more details online at \href{https://geschlechtsneutral.net/}{https://geschlechtsneutral.net}}
It introduces a fourth gender, including new pronouns, articles, and suffixes.
For example, \textit{der Arzt} (\textit{the doctor.MASC.SG}) is changed to \textit{de Arzte} (\textit{the doctor.FEM.MASC.NEUT.SG}).

\section{The \Lou{} Dataset}\label{sec:dataset}
\Lou{} marks the largest collection of reformulated instances for German text classification.
With 3.6k reformulations following six reformulation strategies, \Lou{} enables thoroughly assessing the impact of gender-fair language and the individual strategies across seven classification tasks.
In the following, we discuss the used data (\autoref{subsec:data}) before focusing on the reformulation study (\autoref{subsec:annotation-study}).
%We discuss the used data (\autoref{subsec:data}) and annotation study (\autoref{subsec:annotation-study}).

\subsection{Data}\label{subsec:data}
We start from three German classification datasets: Detox \citep{demus-etal-2022-comprehensive}, GermEval-2021 \citep{risch-etal-2021-overview}, and X-Stance \citep{vamvas2020xstance}. 
We select them since they cover established tasks and minimize the reformulation effort because single instances are annotated with multiple labels. 
For example, Detox provides labels for sentiment analysis, hate-speech, and toxicity detection.
%We selected three German classification datasets because they cover established tasks (like sentiment analysis) and minimize the annotation effort by providing annotated labels for seven classification tasks.
%For example, Detox \citep{demus-etal-2022-comprehensive} annotates every example for sentiment analysis, hate-speech, and toxicity detection.
%Include in CR: In addition, X-Stance \citep{vamvas2020xstance} covers the popular stance detection task and provides the potential to evaluate multi-lingual transfer to German.
Further details and statistics of the datasets are provided in the Appendix \autoref{subsec:app:data}.
\begin{table*}[t]
\centering
    \setlength{\tabcolsep}{2pt}
    \resizebox{1\textwidth}{!}{%
        \begin{tabular}{lcccccccccc}
        \toprule
            % & & & \bf \multicolumn{2}{c}{Overlap} & \bf \multicolumn{3}{c}{Average Train-Test Differences}\\
             \bf Task & \bf Instance & \bf Label\\
        \midrule
      \rowcolor{lightgray}
    \multicolumn{3}{c}{X-Stance \citep{vamvas2020xstance}}\\
    \bf Stance & \textit{\textbf{Topic}: Integration, \textbf{Text}: Integration ist das A und O im Umgang mit Ausländischen \colorbox[HTML]{FFFF05}{\underline{Mitbürger*innen}  \textsubscript{\GenderStern{}}}.} & \texttt{favor} &
 \\
 \textit{Translation} & \textbf{Topic}: Integration, \textbf{Text}: Integration is the be-all and end-all when dealing with foreign \underline{citizens}.
 \\
 
      \rowcolor{lightgray}
    \multicolumn{3}{c}{GermEval-2021 \citep{risch-etal-2021-overview}}\\
    \bf Engaging & \textit{\textbf{Text}: Die Möglichkeit, dass Trump gewinnt ist groß, weil \colorbox[HTML]{FFFF05}{\underline{seine Konkurrenz} \textsubscript{\Neutral{}}} so schwach ist.} & \texttt{engaging}
    \\
    \bf Fact-Claiming & \textit{\textbf{Text}: Die Möglichkeit, dass Trump gewinnt ist groß, weil \colorbox[HTML]{FFFF05}{\underline{ens Gegnere} \textsubscript{\Dee{}}} so schwach ist.} & \texttt{no fact claimed}\\
    \bf Toxicity & \textit{\textbf{Text}: Die Möglichkeit, dass Trump gewinnt ist groß, weil \colorbox[HTML]{FFFF05}{\underline{seine Gegnerin oder Gegner} \textsubscript{\Doppelnennung{}}} so schwach ist.
    } & \texttt{not toxic}\\

 \textit{Translation} & \textbf{Text}: The possibility that Trump will win is high because of his \underline{opponents}. &
 \\
       \rowcolor{lightgray} \multicolumn{3}{c}{Detox \citep{demus-etal-2022-comprehensive}} \\ 
    \bf Hate-Speech & \textit{\textbf{Text}: NRW Lusche ihr seid scheiße nein du bist es! \colorbox[HTML]{ff8989}{\underline{Ein Freund}  \textsubscript{\texttt{Masculine}}} aller Schwulen Spahnferkels.} & \texttt{hate-speech}\\
    \bf Sentiment & \textit{\textbf{Text}: NRW Lusche ihr seid scheiße nein du bist es! \colorbox[HTML]{FFFF05}{\underline{Ein:e Freund:in} \textsubscript{\GenderDoppelpunkt{}}} aller Schwulen Spahnferkels.} & \texttt{negative}\\
    \bf Toxicity & \textit{\textbf{Text}: NRW Lusche ihr seid scheiße nein du bist es! \colorbox[HTML]{FFFF05}{\underline{Ein\_e Freund\_in} \textsubscript{\GenderGap{}}} aller Schwulen Spahnferkels.} & \texttt{toxic} \\
 \textit{Translation} & \textbf{Text}: NRW losers you suck, no you are! A \underline{friend} of all gay Spahn pigs.
 \\
        \bottomrule
        \end{tabular}
    }
    \caption{Example of the seven German classification tasks in \Lou{}, along with their translations. Gender-fair reformulation strategies (subscript) are highlighted in yellow, and masculine formulations are in orange.
    }
    \label{tab:examples}
\end{table*}

\paragraph{X-Stance \citep{vamvas2020xstance}} annotates multi-lingual texts (\textit{de}, \textit{fr}, \textit{it}) with their stance (\textit{favor} or \textit{against}) regarding 12 topics.

\paragraph{GermEval-2021 \citep{risch-etal-2021-overview}} annotates social media texts with three binary properties: toxicity, fact-claiming, and engaging. 

\paragraph{Detox \citep{demus-etal-2022-comprehensive}} annotates social media texts regarding sentiment, hate-speech, and toxicity. 
Following original instructions, we derive classification labels from the provided raw annotations.
Because the additional training data used in the original paper is unavailable, we sub-sample a more label-balanced train set.

\subsection{Reformulation Study}\label{subsec:annotation-study}
For every dataset, we sampled 200 test instances containing at least one gender-specific term, identified via Diversifix\footnote{A tool for gender-fair language (\href{https://diversifix.org/}{https://diversifix.org/}).}.
We employ an iterative approach involving both eight amateurs and two professionals to ensure the quality of gender-fair reformulations. 
While amateurs have an average self-determined moderate experience of using gender-fair language (more details in Appendix \autoref{subsec:app:annotation}), professionals have a linguistic background and use it daily.
Within this study, we ensure \textit{high-quality}, meaning that specific reformulation strategies are correctly applied without grammatical errors, and \textit{consistency} as semantics and annotated task labels of the original instances are preserved.
Therefore, we avoid using large LMs for annotation as they do not produce gender-fair language with sufficient quality \citep{DBLP:conf/eacl/SavoldiPFNB24}.

\begin{table}[t]
\centering
    \setlength{\tabcolsep}{3pt}
    \resizebox{0.48\textwidth}{!}{%
        \begin{tabular}{lcccr}
        \toprule
            % & & & \bf \multicolumn{2}{c}{Overlap} & \bf \multicolumn{3}{c}{Average Train-Test Differences}\\
              &\bf \Doppelnennung{}  & \bf \GenderStern{} & \bf \Neutral{}& \bf Avg.\\
        \midrule
    \textit{X-Stance} & 7.5\% & 10.5\%  & 10.0\% & 9.3\%  \\
    \textit{Germeval-2021} & 21.0\% & 31.0\%  & 21.5\% & 24.5\%\\\midrule
    Avg. & 14.3\% & 20.8\%  & 15.6\% & 16.9\%\\
        \bottomrule
        \end{tabular}
    }
    \caption{Percentage of proofreading corrections compared to amateur reformulations.}
    \label{tab:proofread}
\end{table}

\paragraph{i) Amateur Annotators}
First, we ask each of the eight amateurs to reformulate 50 distinct instances from X-Stance and GermEval-2021 regarding the \Doppelnennung{}, \GenderStern{}, and \Neutral{} strategies, leading to 1.2k distinct reformulations. 
The annotators need to fulfill the reformulation according to a given strategy.
Other grammatical errors should be ignored to ensure the dataset's validity.
We speed up the reformulation with automatic suggestions from Diversifix and highlight the relevant parts (as in \autoref{fig:overview}) and provide examples in the interface.\footnote{Annotation guidelines are available \href{https://github.com/UKPLab/lou-gender-fair-reformulations}{online}.}

\paragraph{ii) Professional Proofreading}
We validate the amateur reformulations with professional proofreading (\textbf{P1}) to ensure \textit{high-quality} of the reformulations.
\autoref{tab:proofread} shows substantial corrections were necessary by \textbf{P1}, \textbf{hinting at the substantially degraded quality of amateur reformulations} (\textbf{RQ1}).
Corrections were necessary in 16.9\% of the cases and up to 31.0\% for \GenderStern{} on the GermEval-2021 data.
At the same time, the nature of the original text matters as GermEval-2021 seems more challenging for amateurs than X-Stance, as its texts are generally longer and less grammatically consistent (social media).
Further categorization of corrections (find details in Appendix \autoref{subsec:app:error}) shows that amateurs particularly struggle, among others, when adapting pronouns (\textit{ein} into \textit{ein*e}) or handling the grammatical number (\textit{Studenten*innen} instead of \textit{Student*innen}).

\paragraph{iii) Proofreading Verification}
Due to the substantial corrections during proofreading, we verify the reliability of \textbf{P1} with another verification round using a subset of 20\% of the instances.
Those were verified by another professional proofreader (\textbf{P2}).
We find a high agreement of 95\% between \textbf{P1} and \textbf{P2}, confirming the reliability of \textbf{P1}.

\paragraph{iv) Detox Dataset and \Dee{} Strategy}
Based on their high reliability, we conduct a fourth iteration with \textbf{P1} including 200 instances from the Detox and the \Dee{} strategy for all three datasets. 

\subsection{Dataset Composition}
Using the reformulations, we compose the final \Lou{} dataset with instances for the seven tasks of X-Stance, GermEval-2021, and Detox (200 ones each).
For all 600 instances, \Lou{} provides reformulations for \Doppelnennung{}, \GenderStern{}, \GenderGap{}, \GenderDoppelpunkt{}, \Neutral{}, and \Dee{} leading to 3.6k distinct reformulations.
Note, we use a regular expression to generate instances for \GenderGap{} automatically and \GenderDoppelpunkt{} based on \GenderStern{} by replacing the star character \textbf{(*)} with a colon \textbf{(:)} or gap \textbf{(\_)}. 
Ultimately, we ensure \textit{consistency} of the reformulations and manually verify a subset of them to ensure that task labels remain valid. 
We find that semantics and the specific task label of instances are unchanged, confirming their validity for analysis (Appendix \autoref{subsec:app:label-verification}).
%Finally, we analyze the task label consistency and find no label flips due to gender-fair reformulations (Appendix \autoref{subsec:app:label-verification}).
\section{Experimental Setup}\label{sec:setup}
The following section outlines the experimental setups used to assess the impact of gender-fair language on classification during inference, including learning paradigms (\autoref{subsec:paradigms}), used encoder and decoder LMs (\autoref{subsec:models}), and evaluation (\autoref{subsec:evaluation}).

\subsection{Learning Paradigm}\label{subsec:paradigms}

\paragraph{Fine-Tuning}
We tune encoder LMs on the original train and dev (without reformulations) set of the seven \Lou{} tasks for five epochs with early stopping. 
We select the best batch size $\{8,16,32\}$ and learning rate $\{5\cdot10^{-5},2\cdot10^{-5},1\cdot10^{-5}\}$ based on the dev performance for every LM and task across three random seeds. 
Then, we tune LMs across ten random seeds to ensure numeric stability. 

\begin{figure*}[t]
    \centering
    \includegraphics[width=1.0\textwidth]{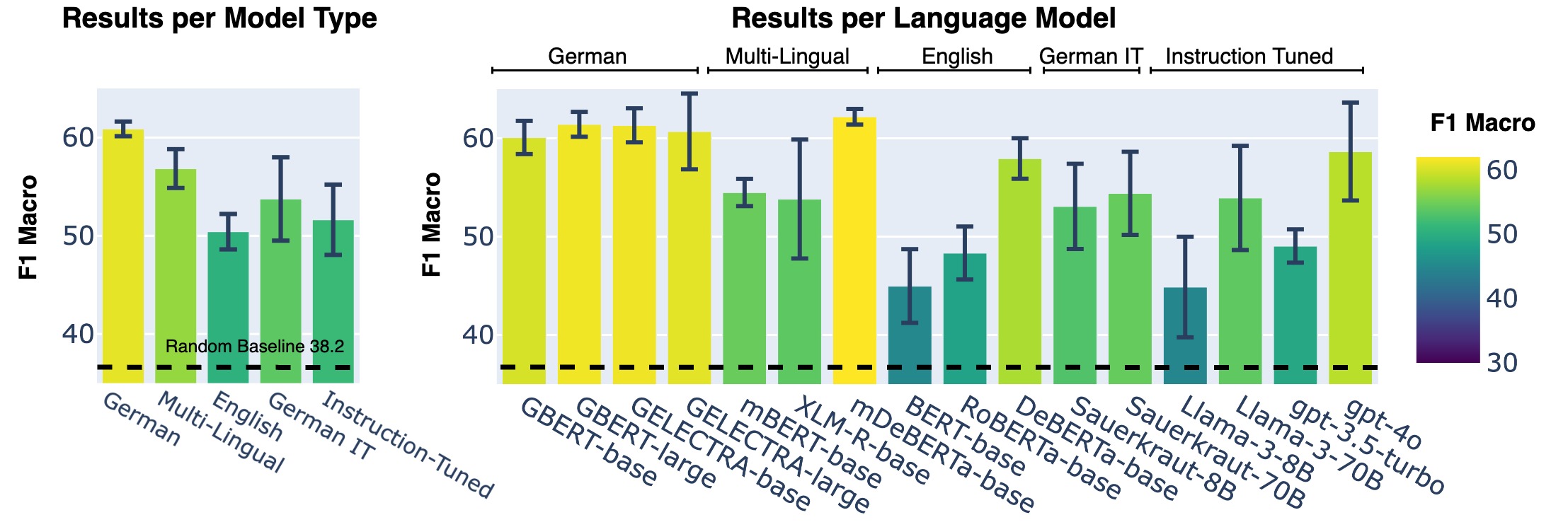}
    \caption{Mean performance and standard deviation, averaged over the seven \Lou{} tasks and seeds (fine-tuning) or prompting templates (ICL) by model type (\textit{left}) or specific LM (\textit{right}).}
    \label{fig:overall_results}
\end{figure*}

\paragraph{In-Context Learning (ICL)}
We evaluate open and closed decoder LMs using their textual response to zero-shot prompts. 
We reuse prompting templates of previous work whenever available and evaluate additional three paraphrased templates to account for variabilities \citep{DBLP:journals/corr/abs-2401-00595}.\footnote{More detail in the Appendix \autoref{subsec:app:icl}} 

\subsection{Models}\label{subsec:models}
We evaluate the following five LM types, including ten encoders and six decoders.\footnote{More details are in Appendix \autoref{subsec:app:lms}} 
Apart from German-specialized and multi-lingual LMs, we further consider English-specialized ones.
As they were mainly trained in English text, they represent the lower bound without a fine-grained understanding of the German language. 
Thus, we assume LMs mainly capture lexical features if English LMs perform competitively. 

\paragraph{German}
We tune four German encoder LMs \citep{chan-etal-2020-germans}: GBERT-base, GBERT-large, GELECTRA-base, and GELECTRA-large.

\paragraph{Multi-Lingual}
We consider three mulit-lingual encoder LMs: mBERT-base \citep{devlin-etal-2019-bert}, XLM-R-base \citep{conneau-etal-2020-unsupervised}, and mDeBERTa-base v3 \citep{DBLP:conf/iclr/HeGC23}.

\paragraph{English}
We evaluate three English LMs: BERT-base \citep{devlin-etal-2019-bert}, RoBERTa-base \citep{DBLP:journals/corr/abs-1907-11692}, and DeBERTa-base v3 \citep{DBLP:conf/iclr/HeGC23}. 

\paragraph{Instruction-Tuned (IT)}
For ICL, we consider four decoder LMs Llama-3-8B and Llama-3-70B \citep{llama3modelcard}, gpt-3.5-turbo, and gpt-4o \citep{DBLP:conf/nips/Ouyang0JAWMZASR22}.

\paragraph{German IT}
In addition, we consider two German specialized large LMs based on Llama-3: Sauerkraut-8B and Sauerkraut-70B.\footnote{Available on \href{https://huggingface.co/VAGOsolutions}{Huggingface}.}

\subsection{Evaluation}\label{subsec:evaluation}
We assess the impact of gender-fair language by comparing predictions on the original test instances with the reformulated ones per LM.
Specifically, we analyze the impact on task level using the $F_1$ macro score and on the instance level by counting prediction flips under gender-fair language. 
We report average and standard deviation across ten random seeds.
We report results on the \Lou{} subset of 200 test instances per task.
Results on these subsets significantly aligned with the full test set, with a Pearson correlation of $\rho=0.97$.

\section{Results}\label{sec:results}
We discuss results obtained across the seven \Lou{} tasks.
%of the seven \Lou{} tasks.
First, we establish our baseline with results on the original samples (i).
Next, we focus on \textbf{RQ2} and the substantial impact of gender-fair language on aggregated evaluation (ii, iii) and individual predictions (iv). 
%We first evaluate the original samples (i) before focusing on \textbf{RQ2} and the substantial impact of gender-fair language on aggregated evaluation (ii, iii) and individual predictions (iv). 
Addressing \textbf{RQ4}, we confirm that existing datasets and evaluations retain their validity under gender-fair language (v). 

\begin{figure*}[t]
    \centering
    \includegraphics[width=1.0\textwidth]{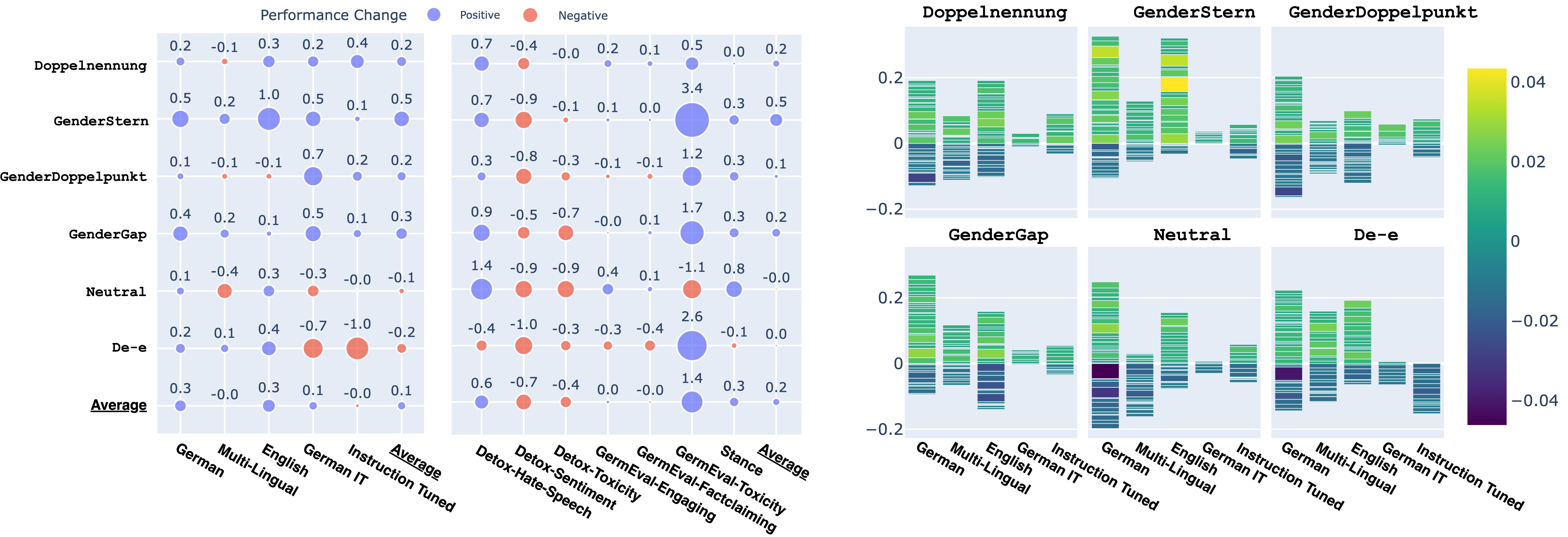}
    \caption{
    Difference between original and reformulated instances for strategies, model types, and tasks in average $F_1$ macro (\textit{left}). The size and the color indicate the difference, whether positive (blue) or negative (red).
    On the \textit{right}, we stack the average difference per LM and seed or prompt template for the model types and strategies.
    }
    \label{fig:diff}
\end{figure*}

\paragraph{i) The value of German specialized LMs.}
\autoref{fig:overall_results} shows the aggregated \Lou{} performance, emphasizing the necessity of specialized German LMs to achieve competitive results. 
On average, German decoders (53.7) outperform general ones by 2.1 points. 
Similarly, German and multilingual encoders (60.9, 56.9) surpass their English counterparts by 10.5 and 6.1 points, respectively.
Notably, \textbf{mDeBERTa demonstrates its practical value for German tasks, marginally outperforming the German-specific encoders}, particularly in challenging scenarios with highly label imbalances like in the \texttt{Detox Hate-Speech} task (see \autoref{tab:dataset_insights} in the Appendix).
The surprisingly strong performance of its English counterpart (DeBERTa) suggests that these LMs may rely more on lexical features than on a nuanced linguistic understanding of the German language.
This assumption is supported by the substantially larger vocabulary sizes of mDeBERTa (250k) and DeBERTa (128k) compared to the 31k tokens of GBERT and GELECTRA.
Interestingly, model size appears less critical, as GBERT-base and GELECTRA-base do not significantly underperform compared to their larger versions.
However, in ICL, model size plays a crucial role, as Llama-3-70B gains 9 points over Llama-3-8B, and GPT-4o 9.7 points over GPT-3.5-turbo.
Interestingly, this trend does not hold for German decoders, where Sauerkraut-8B remains notably competitive with Sauerkraut-70B, showing only a 1.3-point difference.
Finally, we compare decoders using ICL and fine-tuned encoders.
They excel in different types of tasks (see \autoref{tab:detailed_baseline} in the Appendix).
For example, ICL outperforms fine-tuning when datasets embody apparent difficulties, like imbalanced labels in \texttt{Detox Hate-Speech}.
Overall, our results generalize previous findings from English to German: \textbf{specialized encoders outperform decoders \citep{mosbach-etal-2023-shot}, and ICL and fine-tuning are supplementary learning paradigms, as demonstrated in \citet{ood-waldis}}.

\paragraph{ii) Gender-fair language substantially impacts the performance.}%Gender-fair language tends to impact performance positively.}

\autoref{fig:diff} (\textit{left}) focuses on the task-level influence of gender-fair language and shows the average difference between the original performance and the six reformulation strategies.
Surprisingly, reformulations tend to improve measurable performance, especially with inclusive strategies, showing 17 improvements out of 20 cases.
In contrast, neutralization (\Neutral{} and \Dee{}) tends to harm performance on average while only improving the performance in 5 out of 10 cases.
Further, \GenderStern{} provides the most improvement, while \Dee{} exhibits the largest performance degradation. 
Interestingly, while \GenderStern{}, \GenderDoppelpunkt{}, and \GenderGap{} minimally differ from each other (more details in Appendix \autoref{subsec:app:tokenization}), their performance considerably varies. 
This observation suggests that specific special characters (\textbf{*}, \textbf{:}, and \textbf{\_}) semantically differ and shows, again, that LMs rely on lexical features rather than on linguistic specialties of the German language.
Comparing the \Lou{} tasks, Detox ones are more impacted than others, and offensive tasks show more impact compared to Germeval-Engaging, Germeval-Factclaiming, and Stance. 
Specifically, reformulations of GermEval-Toxicity show a significant impact. 
Notably, LMs perform at a lower level on these sensitive tasks, hinting that task difficulty and the impact of gender-fair language are connected.
These insights show that \textbf{even minor changes have big effects, in particular for challenging tasks.}

\paragraph{iii) Aggregation across tasks may hide the impact of gender-fair language.}
We stack in \autoref{fig:diff} (\textit{right)} the differences of every LM and seed or prompting template (German IT and Instruction Tuned) separately.
This detailed analysis shows that the impact of gender-fair language vanishes when aggregating across tasks. 
While the \Neutral{} strategy showed a small impact (-0.1 points $F_1$ macro), the stacked analysis reveals substantial positive and negative effects. 
These insights highlight that \textbf{only a detailed analysis provides the full picture of the impact of gender-fair language.}

\begin{figure}[t]
    \centering
    \includegraphics[width=0.49\textwidth]{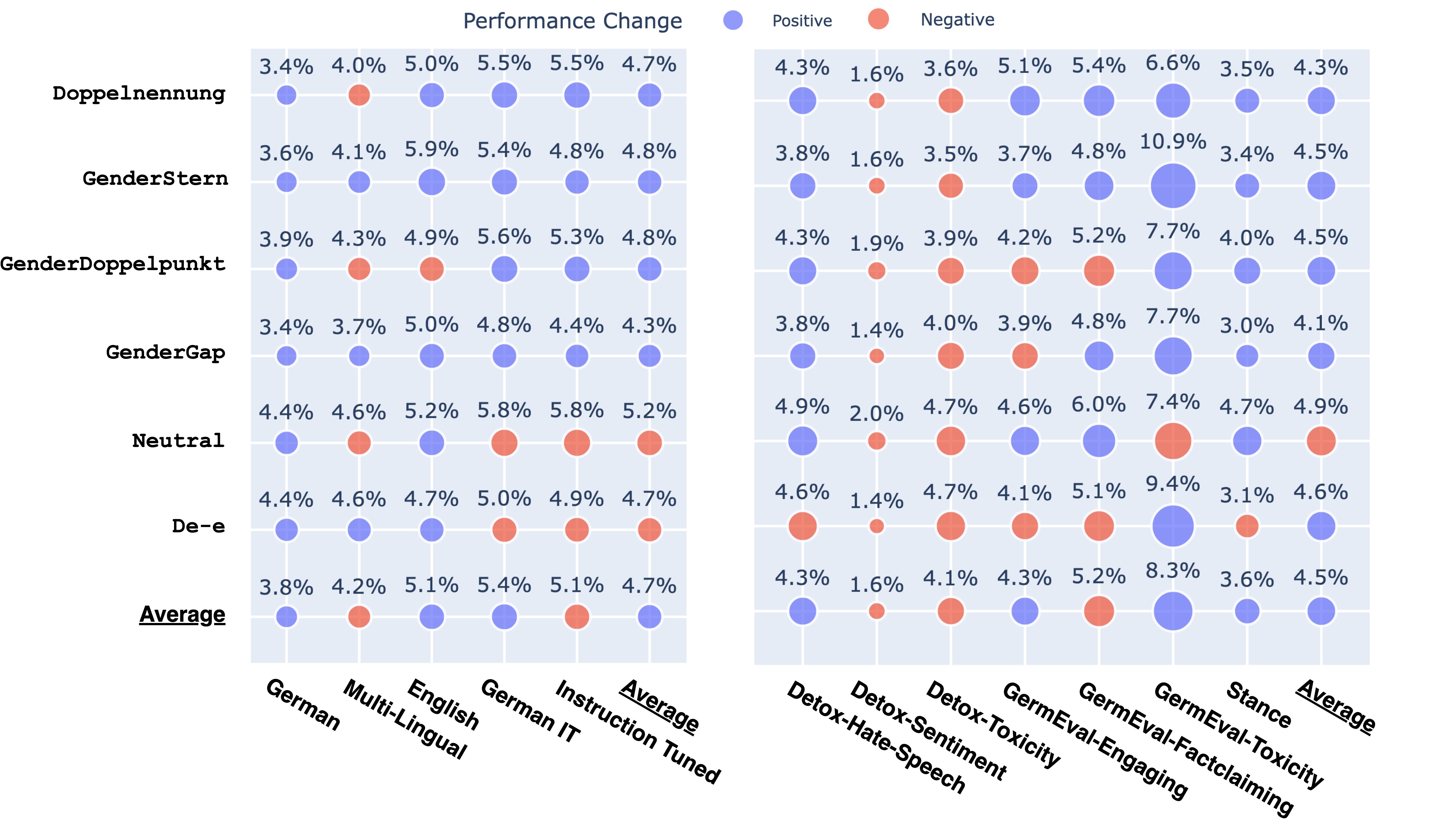}
    \caption{
    Label flip fractions for strategies, model types, and tasks. Size indicates the label flip fraction under gender-fair language and the color positive (blue) or negative (red) effect on aggregated performance.
    }
    \label{fig:flip}
\end{figure}

\begin{figure}[t]
    \centering
    \includegraphics[width=0.48\textwidth]{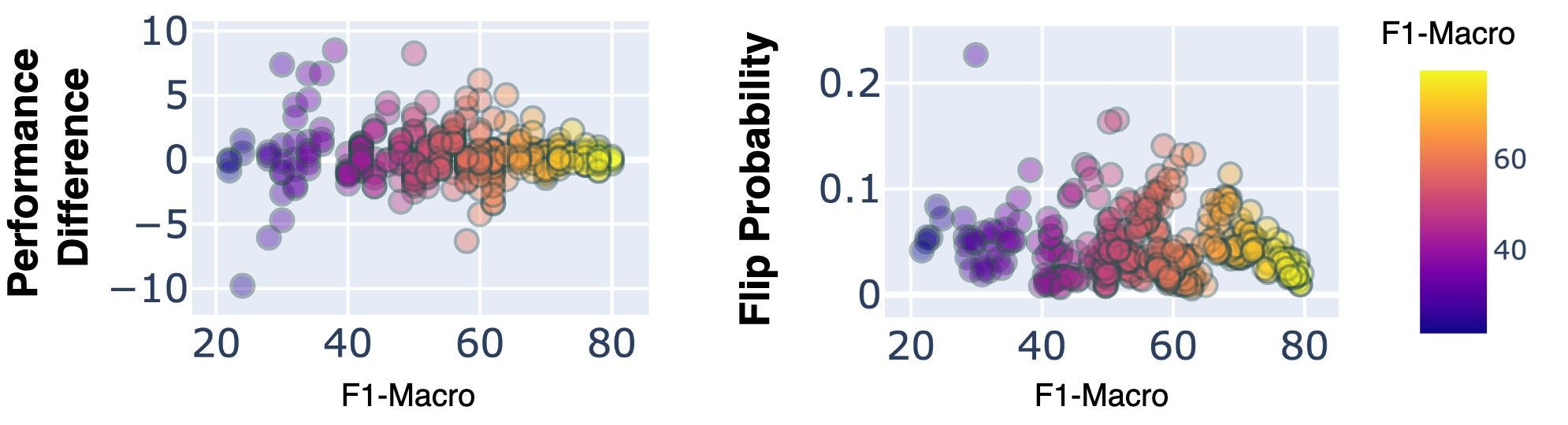}
    \caption{Performance difference and flip fraction against LMs' $F_1$ macros score of each task and strategy.}
    \label{fig:f1_proba_diff}
\end{figure}

\paragraph{iv) Gender-fair language triggers significant label flips.}
We analyze the impact on individual predictions as the fraction of label flips under gender-fair language.
\autoref{fig:flip} shows reformulations flipping labels on average in 4.6\%.
Analyzing the model types (\textit{left}) shows less variability but fewer flips for encoders than decoders, in particular for German specialized ones (German $<$ Multi-Lingual $<$ English).
In contrast, the flip fraction is more spread across tasks (\textit{right}).
While detox-sentiment shows the smallest flip fraction, germinal-toxic exhibits the largest one up to 10.9\% in combination with \GenderDoppelpunkt{}.
Relating to previously discussed results, \GenderGap{} shows, again, a different pattern (less flips) than \GenderStern{} and \GenderDoppelpunkt{} for German, multi-lingual, and English LMs.
This consistent finding demonstrates that even \textbf{minimal syntactic variations of gender-fair language significantly impact single predictions.}
Comparing with the performance differences in \autoref{fig:diff} (\textit{left}) reveals that the label flip fractions provide a different perspective on the impact of gender-fair language. 
These two measures are moderately correlated ($\rho=0.47, p<0.05$) and show substantially different relations to the absolute performance in \autoref{fig:f1_proba_diff}.
However, both measures tend to be less pronounced when LMs perform on a lower level, hinting again at a connection between task difficulty and the impact of gender-fair language. 

\paragraph{v) The consistency of evaluations under gender-fair language.}
We compare the model rankings when evaluating the original or reformulated instances.
%examples 
We find significant ($p<0.05$) high correlations ($\rho \geq 0.95$), meaning that LM rankings are consistent among original and reformulated instances. 
As a result, \textbf{existing datasets retain their validity for evaluations focusing on the supremacy of specific LMs.}
\section{Analysis}\label{sec:analysis}
Focusing on \textbf{RQ3}, we discuss the pronounced effect of reformulations on lower model layers (i) and find reformulations significantly alter attention patterns and decrease prediction certainty (ii), and these properties are crucial for label flips (iii).

\paragraph{i) Gender-fair language affects lower LM layers.}
We analyze how LMs process gender-fair language internally by computing layer-wise average embeddings \citep{reimers-gurevych-2019-sentence} of the original and reformulated text ($s$ and $s'$).
Afterward, we isolate the reformulation within these embeddings as $r=s-s'$.
Then, we test how well we can distinguish the different strategies with $r$ using KMeans \citep{DBLP:journals/tit/Lloyd82} clustering for every layer separately. 
Across all LMs, we find statistically significant (p$<$0.05) negative correlations between the layer numbers and the cluster performance, rand index ($\rho$=-0.40), mutual information ($\rho$=-0.56), completeness ($\rho$=-0.55), and homogeneity ($\rho$=-0.54).
As lower layers account for syntactic information and their degree of contextualization is lower \citep{tenney-etal-2019-bert}, \textbf{gender-fair language has a syntactic impact.}
%todo for CR: Across all layers, we note a moderate match between predicted clusters and reformulation strategy with a rand index of 0.16, mutual information of 0.35, completeness of 0.43, and homogeneity of 0.30. 

\begin{figure}[t]
    \centering
    \includegraphics[width=0.48\textwidth]{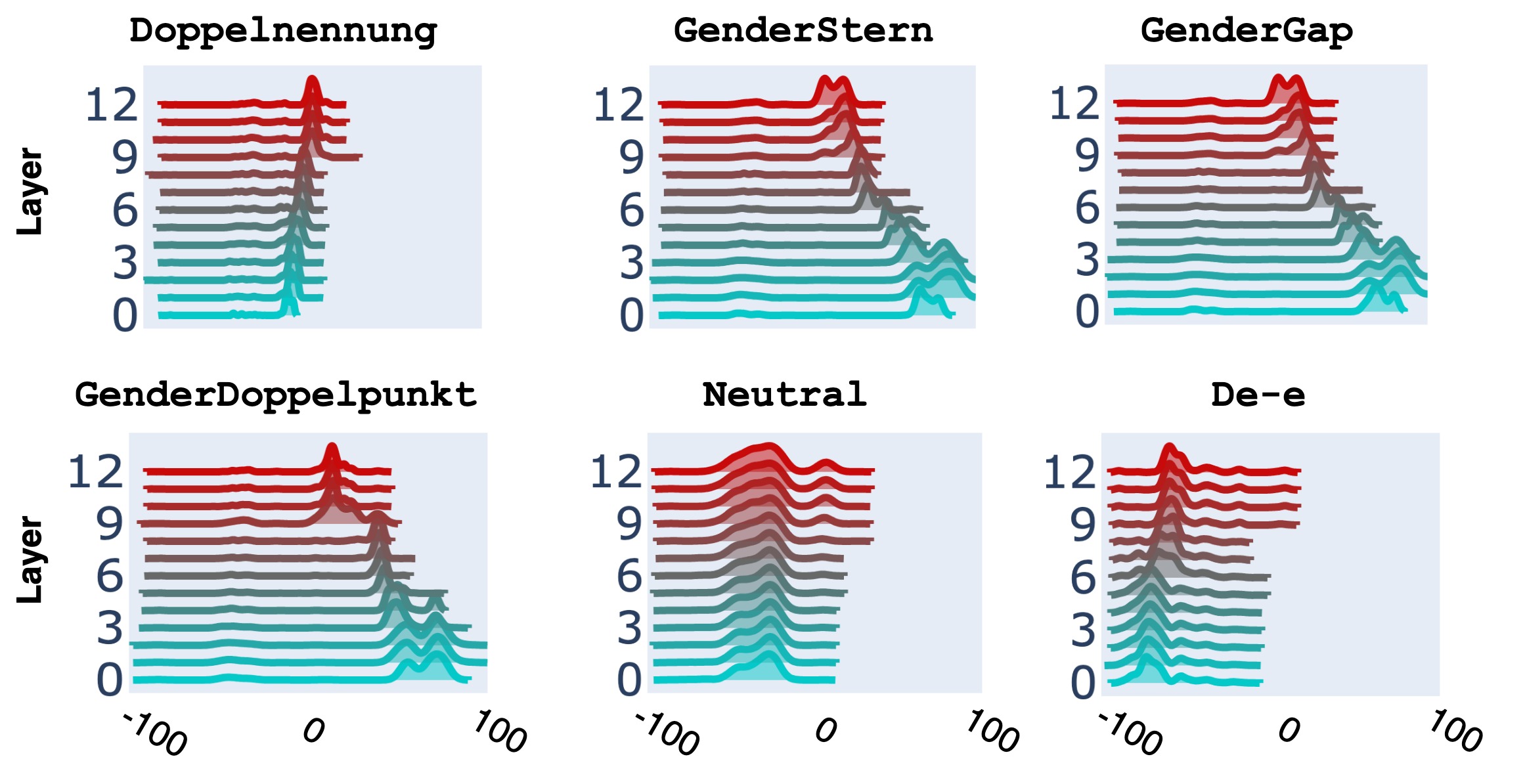}
    \caption{Kernel-density plot of the 1D projected reformulation embeddings $r$ using t-SNE for all six strategies and 13 layers (x-axis) of GBERT-base, including the embedding layer (0).}
    \label{fig:1d_embeddings}
\end{figure}

Next, we qualitatively analyze and show in \autoref{fig:1d_embeddings} that the six strategies are better distinguishable on lower layers by projecting $r$ for all layers to 1D using T-SNE.
While these plots focus on GBERT-base only, we observe similar patterns for other LMs (Appendix \autoref{subsec:app:detail-analysis}).%these plots correspond to GBERT-base, 
\Doppelnennung{}, \Neutral{}, and \Dee{} are more different, while strategies using gender character (\GenderStern{}, \GenderGap{}, and \GenderDoppelpunkt{}) overlap.
Noteworthy, the specific special characters are again crucial, \GenderDoppelpunkt{} (\textbf{:}) differs from \GenderStern{} (\textbf{*}) and \GenderGap{} (\textbf{\_}).
These insights confirm again that the impact of gender-fair language is primarily syntactic. 

\paragraph{ii) Reformulations change instances and how LMs process them.}
Next, we examine the impact of reformulation on input instances and how language models (LMs) process them. 
We focus on surface properties such as normalized instance length and normalized Flesch score (readability, \citet{flesch1948new}), prediction certainty (for encoders only), and LMs’ attention patterns.
We normalize both instance length and Flesch score between zero and one for each task independently. 
Attention patterns are characterized by the maximum attention and its variation (standard deviation) across input tokens. 
Specifically, we analyze how the prediction token attends input tokens: either the classification token (\textit{[CLS]} or \textit{}) for encoders or the first \textit{next-token} for decoders.
To ensure comparability, we exclude tokens affected by the reformulation to ensure consistent attention vector lengths of original and reformulated instances.

\autoref{tab:diff-props} shows that reformulated instances are longer and less readable (lower Flesch). 
These differences are less pronounced for correct predictions, shorter, and more readable than others.
These surface-level changes are known to impact LMs \citep{DBLP:journals/corr/abs-2312-11779}.
Next, LMs show less certainty for reformulated instances, mainly when they cause a label flip (-10.0) or are correct (-2.03).
Consequently, LMs are even less sure when reformulations flip to the correct label (-11.1) and tend to increase attention variation and maximal attention.
This effect is, again, most pronounced for reformulations causing a flip and/or are correct.
These insights show that \textbf{reformulations alter attention patterns and potentially reduce the impact of spurious correlations}, a known drawback for tasks like in \texttt{Hate-Speeech} \citep{attanasio-etal-2022-entropy} or \texttt{stance} detection \citep{thorn-jakobsen-etal-2021-spurious, beck-etal-2023-robust}.

\begin{table}[t]
\centering
    \setlength{\tabcolsep}{10pt}
    \resizebox{0.48\textwidth}{!}{%
        \begin{tabular}{lccccc}
        \toprule
              && \multicolumn{2}{c}{\textbf{Flip}} &\multicolumn{2}{c}{\textbf{Correct}} \\\cmidrule(lr){3-4}\cmidrule(lr){5-6}
              & Overall  & No  & Yes & No  & Yes \\	
        \midrule
        Norm. Flesch & -2.01 & -2.00 & -2.3 & -2.08 & -1.44\\
        Norm. Length & +1.73 & +1.74 & +1.69 & +1.76 & +1.42\\
        Prediction Certainty &  -0.14 & +0.35 & -10.0& +0.49 & -2.03\\\midrule
        Attention Max & +0.41 & +0.37 & +1.14 & +0.33 & +0.56\\
        Attention Variation & +0.10 & +0.08 & +0.23 & +0.07 & +0.13\\\midrule
        \bottomrule
        \end{tabular}
    }
    \caption{Change (stat. sig. at $p<0.05$) between original and reformulated properties, overall, when instances flip or not, or are correct or not.}
    \label{tab:diff-props}
\end{table}

\begin{figure}[t]
    \centering
    \includegraphics[width=0.48\textwidth]{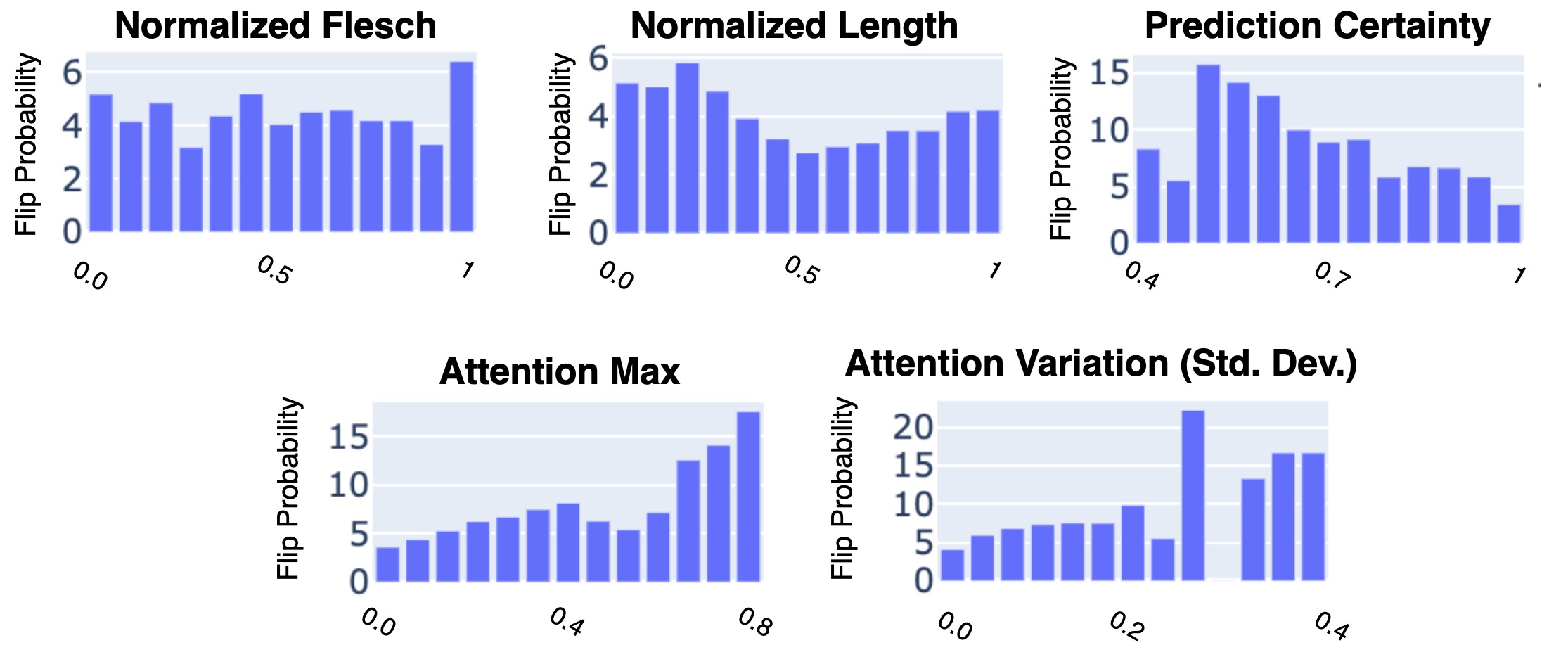}
    \caption{Distribution of instance properties and label flip fractions are statistically significant ($p<0.05$).
    } 
    \label{fig:dist_probabilities}
\end{figure}

\paragraph{iii) The surface properties of instances cause flips.}
\autoref{tab:flip-props} shows that the predicted labels of reformulated instances flip when the original ones are shorter, less readable (lower Flesch), and when LMs show lower prediction certainty.
From higher attention maximum (5.1 vs. 1.4) and variation (1.4 and 0.8) of flipped instances, LMs give higher attention to single tokens, potentially causing a drop in certainty. 
These observations align with our previous results and analyses, which show that the influence of gender-fair language is stronger when the task is difficult, and LMs tend to be less sure.
Specifically, we found an average certainty of 92.4 (no flip) and 89.2 (flip) for GermEval-Toxicity, with a particularly strong impact of gender-fair language. 
\autoref{fig:dist_probabilities} shows the relation between the different properties and the label flip fraction in more detail. 
While the Flesch score shows less pronounced effects, the flip fraction tends to be higher (up to 6\%) for shorter instances.
Further, the flip fraction is crucially higher when an LM predicts with less certainty (up to 15\%) and exhibits a high attention maximum or variation, up to 15\% and 20\%).

\begin{table}[t]
\centering
    \setlength{\tabcolsep}{15pt}
    \resizebox{0.48\textwidth}{!}{%
        \begin{tabular}{lcc}
        \toprule
              & \bf Flip==No & \bf Flip==Yes \\	
        \midrule
        Norm. Flesch & 71.1$\pm$11.4 & 70.7$\pm$11.7\\
        Norm. Length & 27.6$\pm$16.9 &  26.8$\pm$16.8 \\
        Prediction Certainty &  95.7$\pm$9.5 & 90.6$\pm$14.3 \\\midrule
        Attention Max & 10.3$\pm$7.9 & 12.0$\pm$9.6 \\
        Attention Variation & 1.8$\pm$1.6 & 2.1$\pm$1.9 \\%\midrule
       % Diff Attention Max & 2.9$\pm$5.2 & 5.1$\pm$7.6 \\
       % Diff Attention Variation & 0.8$\pm$1.4 & 1.4$\pm$0.2 \\
        \bottomrule
        \end{tabular}
    }
    \caption{Properties of instances when their reformulation causes a label flip or not, including surface properties (Flesch and length), prediction certainty, and attention patterns.
    All differences are stat. sig. ($p<0.05$).}
    \label{tab:flip-props}
\end{table}

\section{Related Work}\label{sec:related-work}
Previous work shows LMs embodying substantially stereotypical bias \citep{kurita-etal-2019-measuring,nadeem-etal-2021-stereoset,DBLP:journals/corr/abs-2206-04615} regarding gender, profession, race, and religion.
In particular, gender bias gained more attention recently \citep{sun-etal-2019-mitigating,gebnlp-2022-gender}.
While some works focus on analyzing the existence of gender bias \citep{zhao-etal-2019-gender,zhao-etal-2020-gender,kaneko-etal-2022-gender}, others aim to reduce this bias \citep{qian-etal-2019-reducing, ravfogel-etal-2020-null,DBLP:journals/corr/abs-2305-13862}.
Another line of work examines the effects of gender-fair language, such as how LMs process (neo)pronouns \citep{brandl-etal-2022-conservative, hossain-etal-2023-misgendered, Gautam2024RobustPU}. 
Gender-fair language is also broadly studied in translation \citep{gitt-2023-gender}, focusing on the effects of gender bias \citep{stanovsky-etal-2019-evaluating}. 
This includes analyzing the acceptance of gender-fair formulations \citep{attanasio-etal-2023-tale}, gender neutralization \citep{piergentili-etal-2023-gender}, the use of interpretability methods \citep{attanasio-etal-2023-tale}, and the impact of pronouns on translation \citep{lauscher-etal-2023-em, amrhein-etal-2023-exploiting}.
Unlike previous work, we focus on the impact of gender-fair language on classification inference. 
Specifically, we present with \Lou{} the first dataset of its kind, to assess the impact of gender-fair language on text classification regarding seven German tasks and analyze this impact in detail. 
\section{Conclusion}\label{sec:discussion}
We comprehensively assess the impact of gender-fair language on German text classification tasks. 
Specifically, we introduce \Lou{}, a high-quality dataset of parallel annotated reformulations that employ various gender-fair strategies. 
Our systematic evaluation and analysis reveal that aggregated evaluations of original data maintain their validity under gender-fair language. 
However, absolute performance tends to increase, while predicted labels can flip with a probability of up to 10.9\%, particularly due to significantly reduced prediction certainty and altered attention patterns.
These findings highlight the importance of considering this linguistic variation, especially since even minor syntactic changes can critically alter how LMs process individual instances. 
Moving forward, we plan to extend \Lou{} to other languages that employ similar gender-fair reformulation strategies, such as Italian and French, and work on adopting LMs for this linguistic variation.

\section*{Limitations}

\paragraph{The Focus on German}
This work solely focuses on gender-fair language in German.
However, we assume our evaluation and analytical pipeline is adaptable to other languages. 
Furthermore, we see empirical insights that the impact is mostly due to syntactic variations of gender-fair language in LMs that can be transferred to another language. 
This is especially plausible since these patterns are consistent across German, multi-lingual, and English LMs. 

\paragraph{Selected Reformulation Strategies}
We select a set of six reformulation strategies to reflect the diversity of options. 
However, we acknowledge the incompleteness as other strategies exist. 
For example, the use of neo-pronouns or the addressing the feminine and masculine gender using the slash character, for example, \textit{Schüler/in}.

\paragraph{Dataset Selection}
The selected German datasets reflect a subset of the available ones.
With them, we aim to cover diverse tasks while optimizing reformulation efforts. 
For example, Detox and GermEval-2021 provide multiple annotations.
However, we do not claim completeness.

\paragraph{Licensing}
For \Lou{}, we adopt the licensing of the underlying datasets and make reformulated instances for X-Stance and GermEval-2021 freely available. 
For Detox, please contact the corresponding author or request the data via the \href{https://tudatalib.ulb.tu-darmstadt.de/handle/tudatalib/4350}{online archive} along with a confirmation of the original dataset access.\footnote{Information are available online under  \href{https://github.com/hdaSprachtechnologie/detox}{https://github.com/hdaSprachtechnologie/detox}}

\section*{Ethical Considerations}
With \Lou{}, we cover a broad selection of German text classification tasks.
This collection includes some datasets with offensive content, like text instances from the GermEval-2021 or the Detox datasets. 
Addressing this issue during reformulation, we collected the explicit consent and willingness to annotate this type of text. 
This includes informing them that potential triggers could arise and that they can stop or skip reformulation without giving reasons when they feel uncomfortable. 
\section*{Acknowledgements}
We thank Vagrant Gautam, Qian Ruan, Tianyu Yang, Nils Dycke, and the anonymous reviews for the helpful feedback, and Mara Haas and Chantal Amrhein for the valuable input on gender-fair writing strategies.
Andreas Waldis has been funded by the Hasler Foundation Grant No. 21024.
The work of Anne Lauscher is funded under the Excellence Strategy of the German Federal Government and States.
\bibliography{latex/anthology, latex/anthology_p2, latex/custom}

\appendix
\section{Appendix}\label{sec:appendix}

\subsection{The Use of AI Assistants}\label{subsec:app:ai-assistants}
We use ChatGPT to rework this paper regarding grammatical correctness and clarity.

\subsection{Additional Information about the Reformulation Study}\label{subsec:app:annotation}

\paragraph{Annotators}
During our iterative annotation study, we distinguish between eight amateurs and two professional (\textbf{P1} and \textbf{P2}) annotators. 
The amateur annotators do not have a linguistic background but are native German-speaking,
They determine their experience in applying gender-fair language from 1 (no experience) to 5 (professional experience) with low (2) to advanced (4) and an average moderate (3).
The professional proofreaders have both a linguistic background as they studied the German language (\textit{Germanistik}, \textit{German studies}) and work in proofreading (\textbf{P1}) or in text agency (\textbf{P2}).

\paragraph{Payment}
The amateur annotators did not receive a payment as they conducted the annotations voluntarily.
In contrast, we pay the principal professional annotator (\textbf{P1}) an hourly rate of 56\$ and the second one (\textbf{P2}) 167\$.

\subsection{Error Analysis of Amateur Annotators}\label{subsec:app:error}
As we found substantial difficulties for amateur annotators in applying gender-fair language with sufficient quality, we analyzed these errors in more detail.
Specifically, the principal professional annotator (\textbf{P1}) categorised the errors regarding seven categories:

\paragraph{1. Personfication} When it is clear that a gender-specific phrase corresponds to a person with a specific gender, gender-fair language is not applicable.
For example, \text{Präsident} (English \textit{president.MASC.SG}) when it is clear that the text refers to Donald Trump.

\paragraph{2. Neutral Substantive} When gender-fair reformulation is unnecessary because the substantive is neutral, like \textit{Gäste} (English \textit{customers.NEUT.PL}).

\paragraph{3. Numerus} Inconsistency in singular and plural in the reformulation. For example, the phrase \textit{die Künstlerin oder den Künstler} should be in plural \textit{die Künstlerinnen oder Künstler} (\textit{the artist.FEM.SG or the artist.MASC.PL}).

\paragraph{4. And/Or} The use of \textit{oder} (English \textit{or}) instead of \textit{und} (English \textit{and}) in \Doppelnennung{}, as \textit{und} is more inclusive an appropriate at this point. 

\paragraph{5. Pronoun} If pronouns were not changed accordingly. For example, \textit{keiner} (English \textit{nobody}) needs to be reformulated into \textit{keine*r} for the strategy \GenderStern{}.

\paragraph{6. Compounds} Errors in compounded words like \textit{Zuschauerreaktionen} (English \textit{audience reactions}). This should be reformulated as \textit{Zuschauer*innenreaktionen} considering the \GenderStern{} strategy.

\paragraph{7. Word root} Errors in the word's root form. For example, when considering \GenderStern{} \textit{Experte*in} (\textit{expert.NEUT.SG}) is not correct, it has to be \textit{Expert*in}. 

\paragraph{8. Other} A collection of other errors.
For example, overlooked reformulations, less common neutral formulations like \textit{Deutsche Personen aus Regierungskreisen} instead of \textit{Deutsche Regierende} (\textit{german person from the government}), or other grammatical errors.

\begin{figure}[t]
 \centering
 \includegraphics[width=0.45\textwidth]{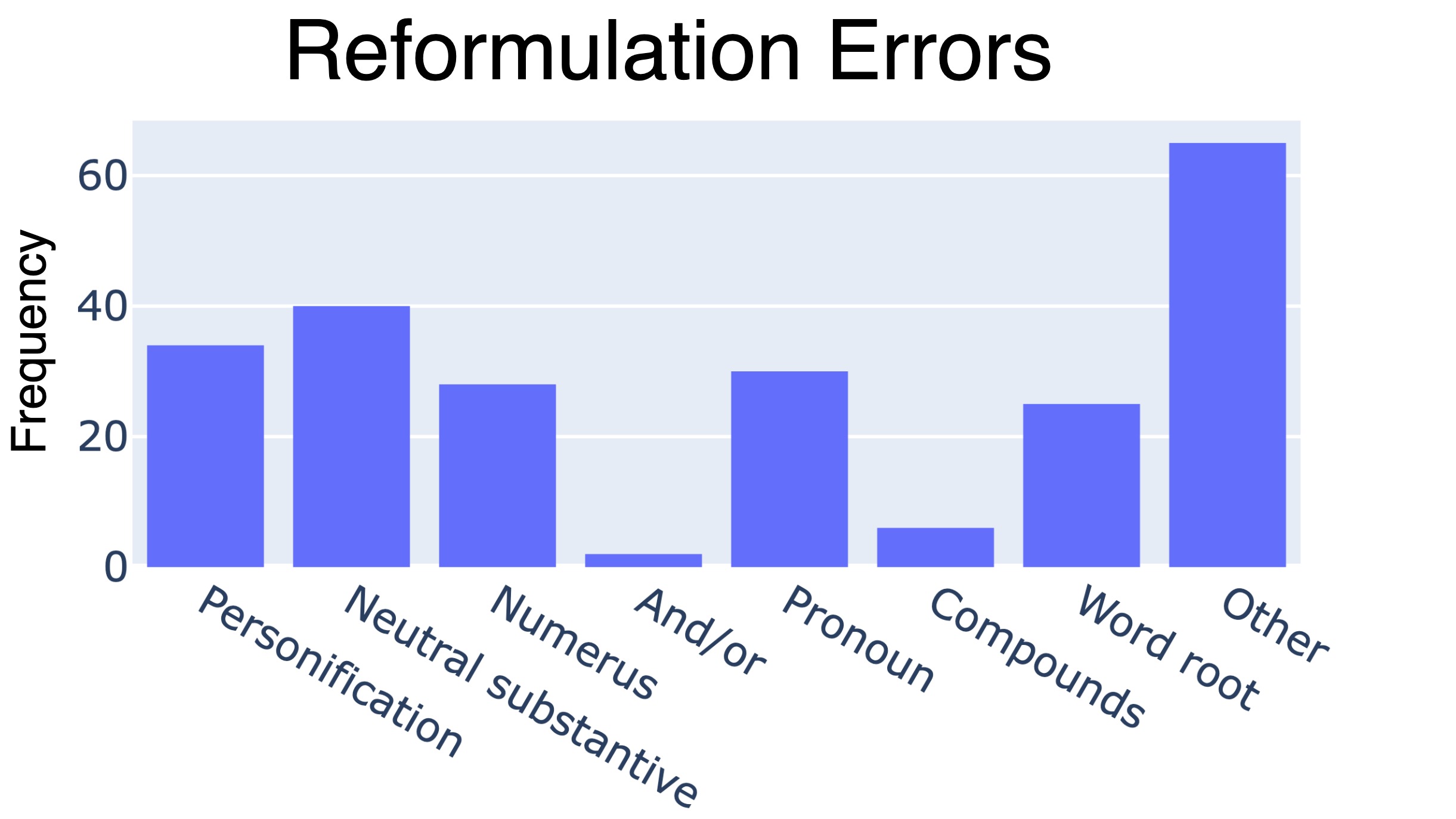}
 \caption{Overview of the categorization frequency when analyzing the errors of the amateur annotators.}
 \label{fig:reformulation_errors}
\end{figure}

\begin{figure*}[t]
 \centering
 \includegraphics[width=1\textwidth]{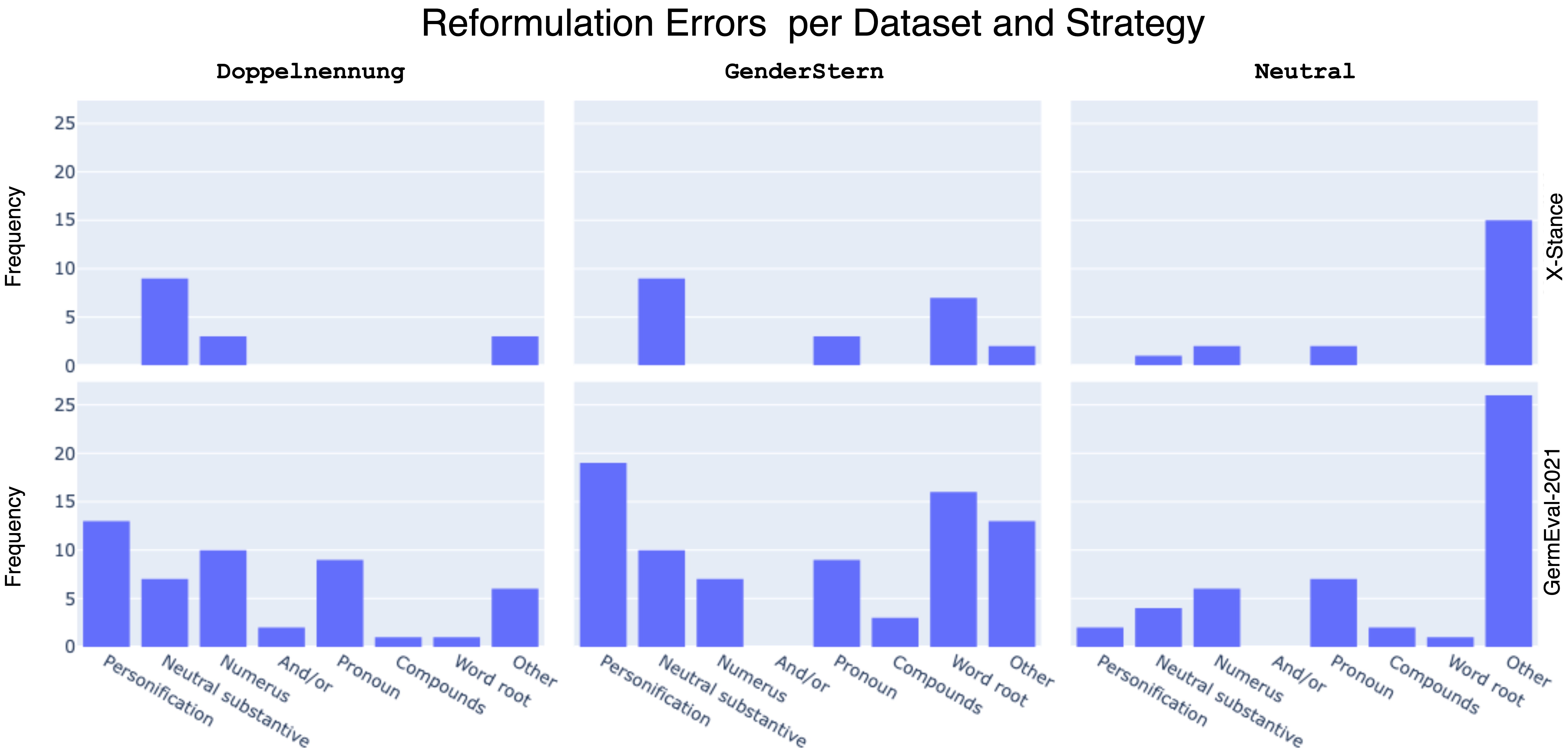}
 \caption{Detailed overview of the categorization frequency when analyzing the errors of the amateur annotators per dataset (row) and strategy (rows).}
 \label{fig:reformulation_errors_detailed}
\end{figure*}

We show in \autoref{fig:reformulation_errors} and \autoref{fig:reformulation_errors_detailed} the frequency of these categories aggregated and per dataset and strategy.
Despite \textit{And/or} and \textit{Compounds}, all categories have a similar frequency.
\textit{Other} is the most frequent category, summarizing many errors. 
However, it is particularly frequent for \Neutral{}, where amateur often used over-complicated and non-usual neutral formulations. 
Regarding the dataset, amateur struggled more with the longer text from GermEval-2021, which often included grammatical errors. 
Concerning the different strategies, \GenderStern{} seems to cause the most errors. 
Particularly prominent are \textit{Personification}, \textit{Neutral Substantive}, \textit{Numerus}, \textit{Pronoun}, and \textit{Word Root}.
These errors show that people struggle to consistently apply gender-fair language, highlighting the need for standardization for broad establishment. 
As their frequency heavily depends on the type and complexity of the text, our insights suggest enrolling in more sophisticated tutoring when solely relying on amateur annotations. 
While being more costly, professional annotators show a clear advantage in providing high-quality reformulations.

\subsection{Additional Information about the Data}\label{subsec:app:data}

\autoref{tab:dataset_insights} show additional information about the considered datasets in \Lou{}.
This includes the average number of tokens (length) and readability (Flesch score), the number of samples per dataset, label distribution, and how train and test label distribution agree using KL divergence. 

\begin{table*}[t]
\centering
 \setlength{\tabcolsep}{3pt}
 \resizebox{1\textwidth}{!}{%
  \begin{tabular}{lcccccccccc}
  \toprule
   % & & & \bf \multicolumn{2}{c}{Overlap} & \bf \multicolumn{3}{c}{Average Train-Test Differences}\\
   &&\multicolumn{3}{c}{\textbf{GermEval-2021}} & \multicolumn{3}{c}{\textbf{Detox}}\\ \cmidrule(lr){3-5} \cmidrule(lr){6-8}
  & \texttt{Stance} & \texttt{Engaging}&  \texttt{Fact-Claiming} & \texttt{Toxicity} & \texttt{Hate-Speech} & \texttt{Sentiment} & \texttt{Toxicity} \\  \midrule
  \bf Length  & 24.5$\pm$15.6 & \multicolumn{3}{c}{30.6$\pm$41.6}  & 28.4$\pm$12.7  &  27.8$\pm$12.8  &  28.6$\pm$12.7 \\  
  \bf Flesch  & 46.3$\pm$9.6 & \multicolumn{3}{c}{59.9$\pm$103.9}   & 50.2$\pm$21.9   &  49.9$\pm$22.7 &  50.3$\pm$21.9 \\  
  \bf \# Samples  & 36,921  & \multicolumn{3}{c}{ 2,276}  & 4,100  & 8,348 & 4,035\\   \midrule
  \bf \multirow{3}{*}{Labels} 
  & \texttt{favor}: 18,227
  & \texttt{yes}: 635
  & \texttt{yes}: 809
  & \texttt{yes}: 801
  & \texttt{yes}: 928
  & \texttt{negative}: 6,253
  & \texttt{yes}: 808 \\
  & \texttt{against}: 18,694
  & \texttt{no}: 1,641
  & \texttt{no}: 1,467
  & \texttt{no}: 1,475
  & \texttt{no}: 3,172
  & \texttt{neutral}: 1,737
  & \texttt{no}: 3,227 \\
  & 
  & 
  & 
  & 
  & 
  & \texttt{positive}: 358
  & \\  
  \bf \# KL Div. & 0.01 & 0.25 & 0.03 & 0.08 & 0.06 & 0.04 & 0.04 \\
  \bottomrule
  \end{tabular}
}
 \caption{Further insights about the seven \Lou{} tasks, including surface properties, size, and label distribution.}
 \label{tab:dataset_insights}
\end{table*}

\subsection{The Effect of Tokenization for Gender-Fair Language}\label{subsec:app:tokenization}

\autoref{tab:tokens} shows how many additional tokens the different reformulation strategies add to the input sentence regarding the various strategies and LMs. 
From these results, \Doppelnennung{} adds the most tokens (6.6 on average), \Dee{} the least with on average 1.6 more tokens, and the other between 3.4 (\GenderStern{}) to 3.1 tokens (\Neutral{}).
Noteworthy, we see apparent differences between \GenderStern{} \& \GenderDoppelpunkt{} and \GenderGap{} for decoder LMs, hinting at the different semantic meanings of these special characters. 
Regarding the LM difference, we note that within and across the model type, LMs with a more extensive vocabulary size tend to add fewer tokens than those with a smaller number of distinct tokens.  
Comparing the results, we do not find a clear correlation between the additional number of tokens and the impact of gender-fair language on an aggregated level or for individual predictions.

\begin{table*}[t]
\centering
 \setlength{\tabcolsep}{3pt}
 \resizebox{1\textwidth}{!}{%
\begin{tabular}{lcccccccc}
\toprule
 & \bf Vocab. Size & \bf \Doppelnennung{} & \bf \GenderStern{}& \bf \GenderDoppelpunkt{} & \bf \GenderGap{} & \bf \Neutral{} & \bf \Dee{} & \bf Average \\
\midrule
GBERT &  31k  &  5.3 &  3.5 & 3.5 & 3.5 &   2.7 &  1.8 &   3.4 \\
GELECTRA   &  31k  &   5.3 &  3.5 & 3.5 & 3.5 &   2.7 &  1.8 &   3.4 \\\midrule
mBERT & 120k &   6.2 &  3.5 & 3.5 & 3.5 &   2.9 &  1.7 &   3.6 \\
XLM-R & 250k &   6.0 &  3.5 & 3.5 & 3.5 &   2.6 &  1.5 &   3.4 \\
mDeBERTa-v3-base   & 250k   &   6.0 &  3.4 & 3.4 & 3.4 &   2.4 &  1.1 &   3.3 \\\midrule
BERT  & 31k &   8.3 &  4.4 & 4.4 & 4.4 &   4.2 &  1.5 &   4.5 \\
RoBERTa  & 50k &   9.9 &  4.4 & 4.4 & 4.4 &   4.6 &  1.7 &   4.9 \\
DeBERTa-v3-base & 128k  &   6.9 &  3.4 & 3.4 & 3.4 &   3.2 &  1.2 &   3.6 \\\midrule
Sauerkraut   & 128k  &   7.1 &  3.2 & 3.2 & 2.7 &   3.2 &  1.6 &   3.5 \\\midrule
Llama-3  & 128k  &   7.1 &  3.2 & 3.2 & 2.7 &   3.2 &  1.6 &   3.5 \\
gpt-3.5-turbo   & 100k  &   7.2 &  3.2 & 3.2 & 2.7 &   3.2 &  1.6 &   3.5 \\
gpt-4o & 250k   &   6.1 &  2.3 & 2.2 & 2.8 &   2.7 &  1.5 &   2.9 \\\midrule\midrule
Average     &   &   6.6 &  3.4 & 3.4 & 3.3 &   3.1 &  1.6 &   3.6 \\

\bottomrule
\end{tabular}
 }
 \caption{Number of additional tokens when comparing the reformulated examples with the original ones. Average across all tasks regarding models and strategies. }
 \label{tab:tokens}
\end{table*}

\subsection{Additional Details In-Context Learning}\label{subsec:app:icl}

Similar to using random seeds when fine-tuning LMs, we use four different prompts to measure the LMs' task capabilities thoroughly. 
Following, we provide examples of these templates for the fact-claiming task. 
For the first template, we follow the previous when task prompts are available and translate them into German, such as hate-speech, toxicity, or stance detection in \citet{beck-etal-2024-sensitivity}.
Composing templates two and three, we rephrase the task instructions. 
For the fourth template, we restructure the prompt and embed the example within the task instructions.

\paragraph{Prompt Template 1}\label{para:template-1}
Geben ist der folgenden Satz, wird in diesem Tatsachen behauptet oder nicht?
Mögliche Antworten sind 'ja', falls im Satz Tatsachen behauptet werden oder ‘nein' falls nicht.
Antworte nur mit einem dieser Möglichkeiten und ohne Erklärung!\\
Text: Die Möglichkeit, dass Trump gewinnt ist groß, weil seine Gegner*innen so schwach ist.
Tatsachen erwähnt: \textbf{nein}

\paragraph{Prompt Template 2}
Die Aufgabe ist es zu erkennen ob im folgenden Satz Tatsachen behauptet werden oder nicht.
Mögliche Antworten sind 'ja', falls im Satz Tatsachen behauptet werden oder ‘nein' falls nicht.
Antworte nur mit einem dieser Möglichkeiten und ohne Erklärung!\\
Text: Die Möglichkeit, dass Trump gewinnt ist groß, weil ens Gegnere so schwach ist.
Tatsachen erwähnt: \textbf{nein}

\paragraph{Prompt Template 3}
Betrachten wir den folgenden Satz. Wird in diesem Tatsachen behauptet?
Antworte mit einem Wort: 'ja' falls der Satz Tatsachen behauptet oder 'nein' falls nicht.\\
Text: Die Möglichkeit, dass Trump gewinnt ist groß, weil seine Konkurreenz so schwach ist.
Tatsachen erwähnt: \textbf{nein}

\paragraph{Prompt Template 4}
Werden im Satz "Die Möglichkeit, dass Trump gewinnt ist groß, weil seine Gegner so schwach ist." tatsachen behauptet oder nicht?
Antworte mit "ja", falls der Satz toxisch ist oder "nein" falls nicht.\\
\textbf{nein}

\subsection{Used Language Models}\label{subsec:app:lms}
We run all of our experiments using Nvidia RTX A6000 GPUs.
Every GPU provides 48GB of memory and 10752 CUDA Cores.
We use the following models from the huggingface model hub:

\begin{itemize}
 \item \href{https://huggingface.co/deepset/gbert-base}{\texttt{deepset/gbert-base}}
 \item \href{https://huggingface.co/deepset/gbert-large}{\texttt{deepset/gbert-large}}
 \item \href{https://huggingface.co/deepset/gelectra-base}{\texttt{deepset/gelectra-base}}
 \item \href{https://huggingface.co/deepset/gelectra-large}{\texttt{deepset/gelectra-large}}
 \item \href{https://huggingface.co/bert-base-multilingual-cased}{\texttt{bert-base-multilingual-cased}}
 \item \href{https://huggingface.co/FacebookAI/xlm-roberta-base}{\texttt{FacebookAI/xlm-roberta-base}}
 \item \href{https://huggingface.co/microsoft/mdeberta-v3-base}{\texttt{microsoft/mdeberta-v3-base}}
 \item \href{https://huggingface.co/bert-base-uncased}{\texttt{bert-base-uncased}}
 \item \href{https://huggingface.co/roberta-base}{\texttt{roberta-base}}
 \item \href{https://huggingface.co/microsoft/deberta-v3-base}{\texttt{microsoft/deberta-v3-base}}
 \item \href{https://huggingface.co/TechxGenus/Meta-Llama-3-70B-Instruct-AWQ}{\texttt{TechxGenus/Meta-Llama-3-70B\-Instruct-AWQ}}
 \item \href{https://huggingface.co/TechxGenus/Meta-Llama-3-8B-Instruct-AWQ}{\texttt{TechxGenus/Meta-Llama-3-8B\-Instruct-AWQ}}
 \item \href{https://huggingface.co/mayflowergmbh/Llama-3-SauerkrautLM-8b-Instruct-AWQ}{\texttt{mayflowergmbh/Llama-3\-SauerkrautLM-8b-Instruct-AWQ}}
 \item \href{https://huggingface.co/tresiwalde/Llama-3-SauerkrautLM-70b-Instruct-AWQ}{\texttt{tresiwalde/Llama-3\-SauerkrautLM-70b-Instruct-AWQ}}
\end{itemize}

\subsection{Detailed Results}\label{subsec:app:detail-results}

\begin{figure*}[t]
 \centering
 \includegraphics[width=1\textwidth]{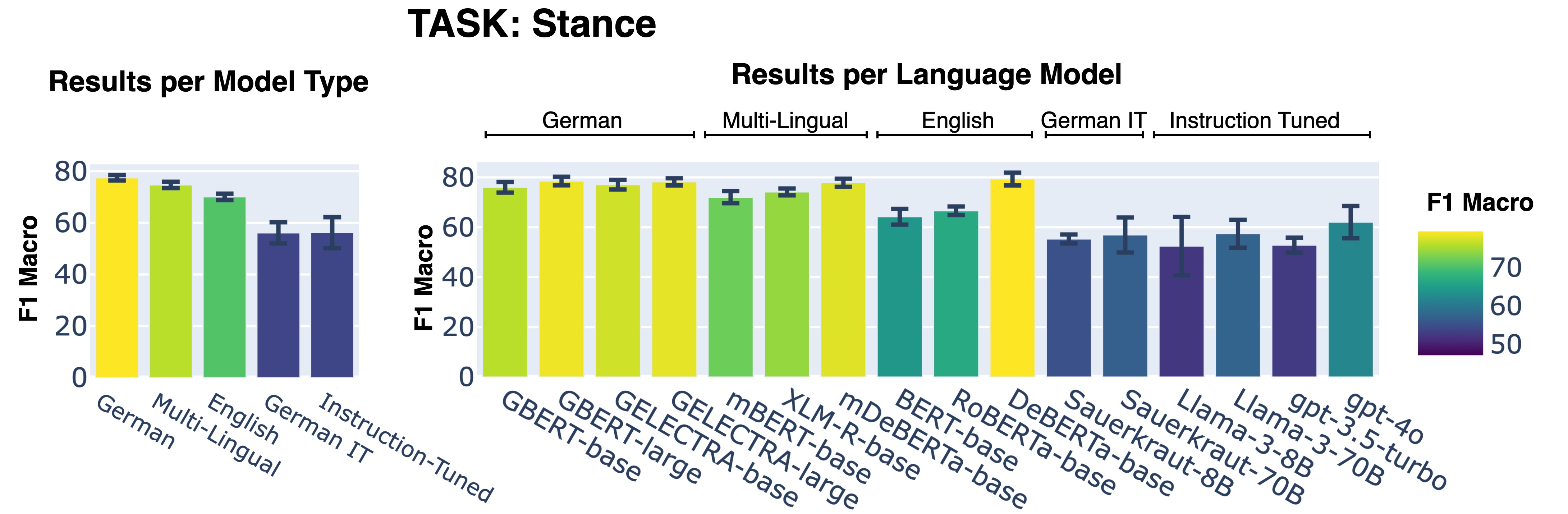}
 \caption{Mean performance and standard deviation for the \texttt{stance} task averaged over and seeds (fine-tuning) or prompting templates (ICL) by model type (\textit{left}) or specific LM (\textit{right}).}
 \label{fig:stance}
\end{figure*}

\begin{figure*}[t]
 \centering
 \includegraphics[width=1\textwidth]{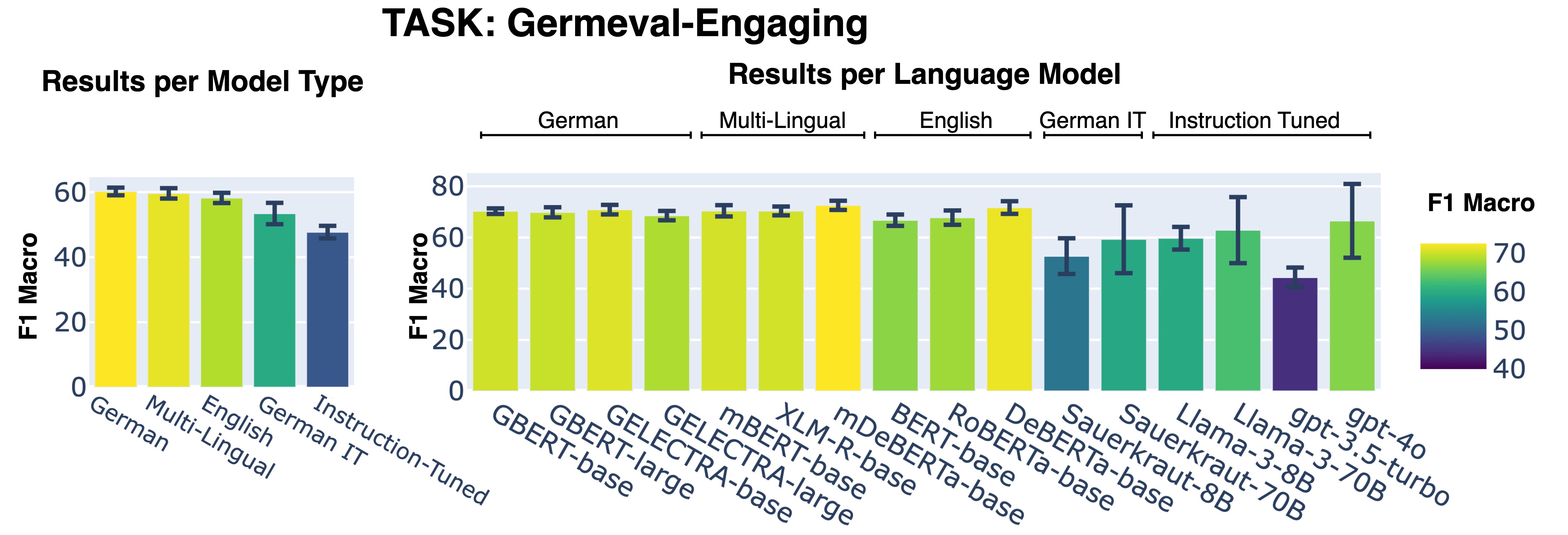}
 \caption{Mean performance and standard deviation for the \texttt{GermEval-Engaging} task averaged over and seeds (fine-tuning) or prompting templates (ICL) by model type (\textit{left}) or specific LM (\textit{right}).}
 \label{fig:germeval-engaging}
\end{figure*}

\begin{figure*}[t]
 \centering
 \includegraphics[width=1\textwidth]{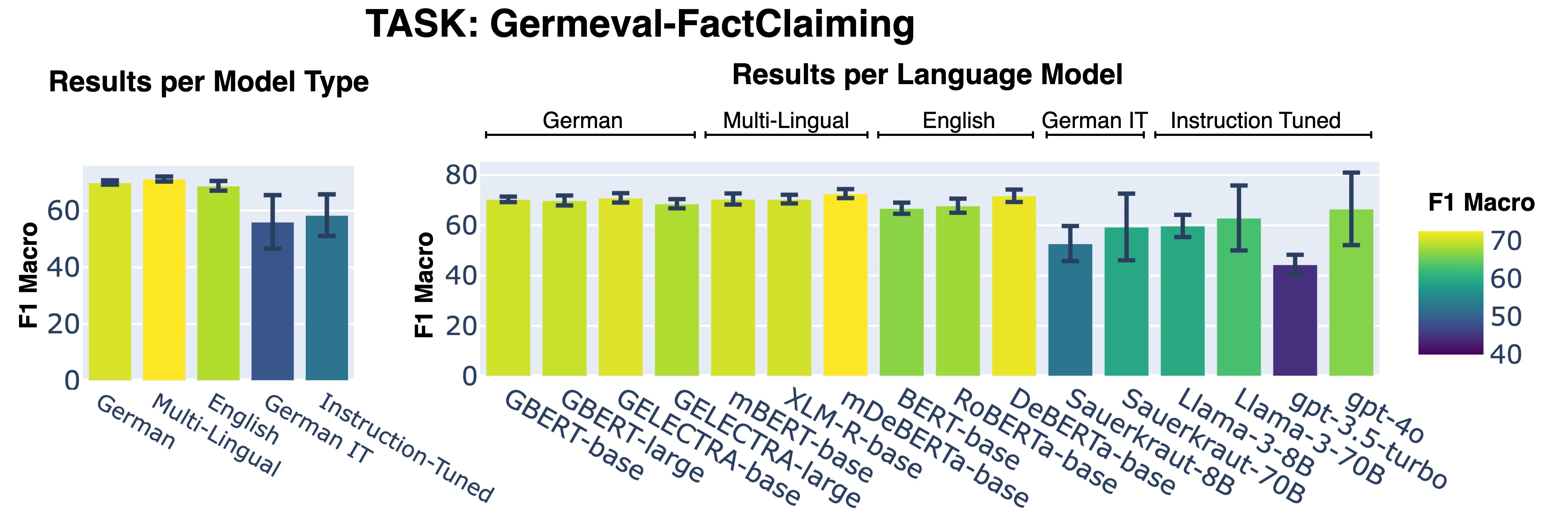}
 \caption{Mean performance and standard deviation for the \texttt{GermEval-Fact-Claiming} task averaged over and seeds (fine-tuning) or prompting templates (ICL) by model type (\textit{left}) or specific LM (\textit{right}).}
 \label{fig:germeval-fact}
\end{figure*}

\begin{figure*}[t]
 \centering
 \includegraphics[width=1\textwidth]{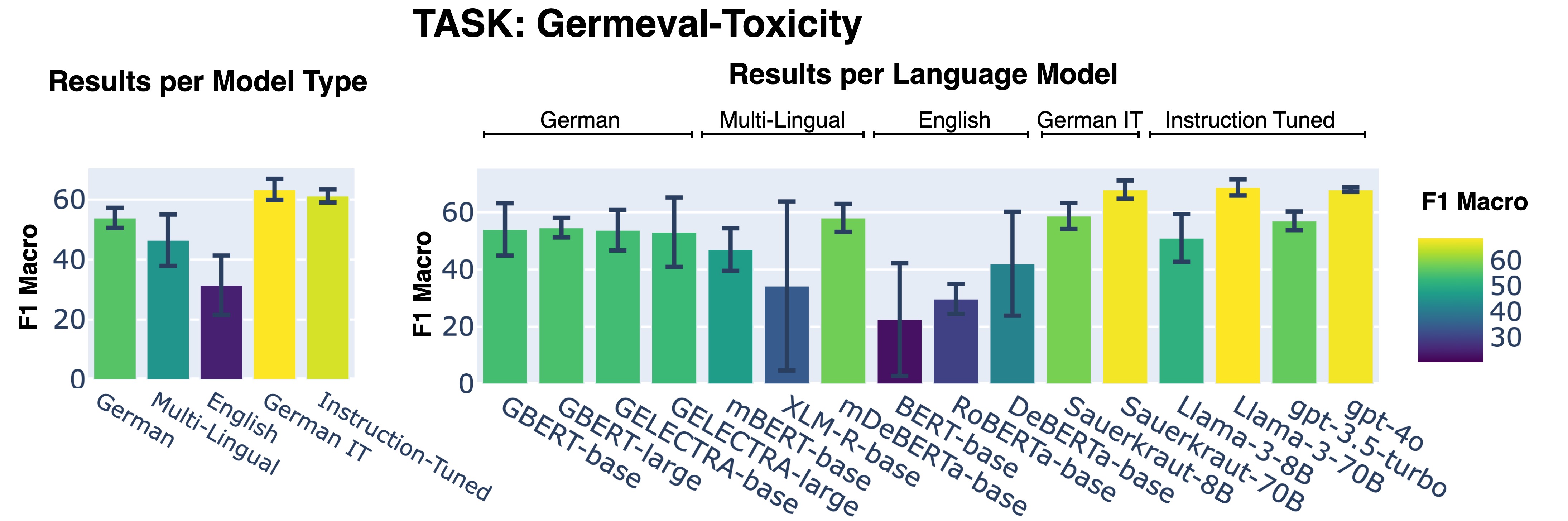}
 \caption{Mean performance and standard deviation for the \texttt{GermEval-Toxicity} task averaged over and seeds (fine-tuning) or prompting templates (ICL) by model type (\textit{left}) or specific LM (\textit{right}).}
 \label{fig:germeval-toxic}
\end{figure*}

\begin{figure*}[t]
 \centering
 \includegraphics[width=1\textwidth]{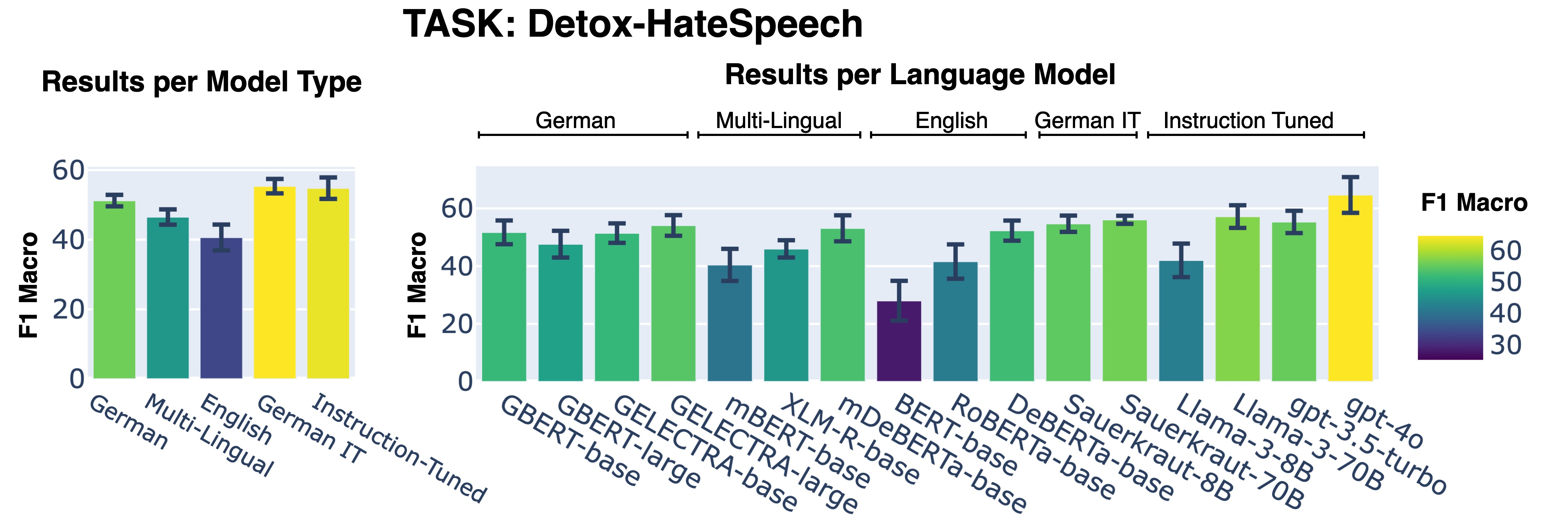}
 \caption{Mean performance and standard deviation for the \texttt{Detox-Hate-Speech} task averaged over and seeds (fine-tuning) or prompting templates (ICL) by model type (\textit{left}) or specific LM (\textit{right}).}
 \label{fig:detox-hatespeech}
\end{figure*}

\begin{figure*}[t]
 \centering
 \includegraphics[width=1\textwidth]{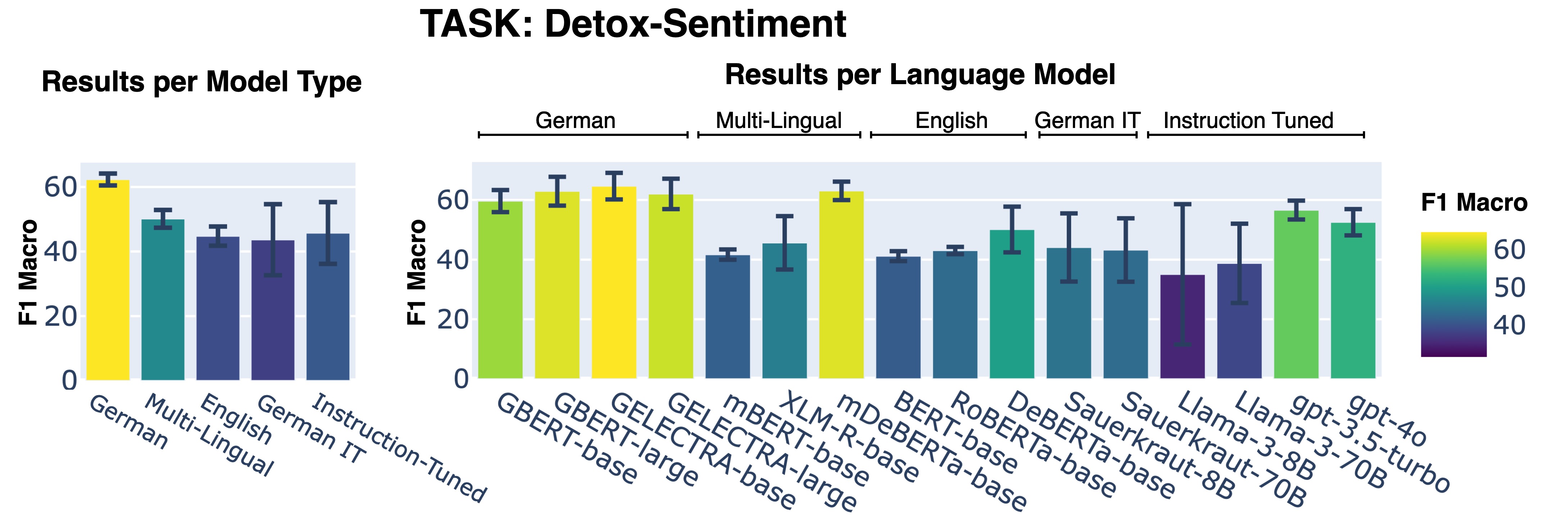}
 \caption{Mean performance and standard deviation for the \texttt{Detox-Sentiment} task averaged over and seeds (fine-tuning) or prompting templates (ICL) by model type (\textit{left}) or specific LM (\textit{right}).}
 \label{fig:detox-sentiment}
\end{figure*}

\begin{figure*}[t]
 \centering
 \includegraphics[width=1\textwidth]{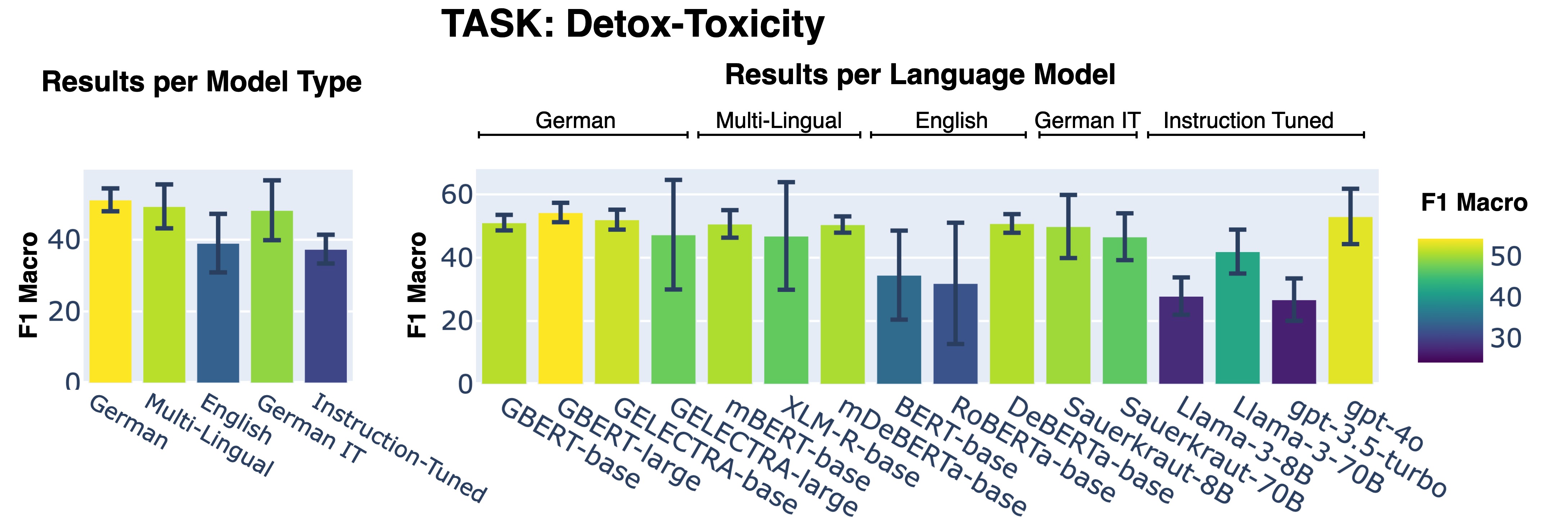}
 \caption{Mean performance and standard deviation for the \texttt{Detox-Toxicity} task averaged over and seeds (fine-tuning) or prompting templates (ICL) by model type (\textit{left}) or specific LM (\textit{right}).}
 \label{fig:detox-toxic}
\end{figure*}

\autoref{tab:detailed_baseline}, \autoref{fig:stance}, \autoref{fig:germeval-engaging}, \autoref{fig:germeval-fact}, \autoref{fig:germeval-toxic}, \autoref{fig:detox-hatespeech}, \autoref{fig:detox-sentiment}, and \autoref{fig:detox-toxic} shows the detailed baseline results covering all the seven \Lou{} tasks for the 16 considered results.
Note that we evaluated the original examples without reformulations.

\begin{table*}[t]
\centering
 \setlength{\tabcolsep}{3pt}
 \resizebox{1\textwidth}{!}{%
\begin{tabular}{lcccccccc}
 \toprule
  &  \multicolumn{3}{c}{\textbf{Detox}} & \multicolumn{3}{c}{\textbf{GermEval-2021}} & & \\ \cmidrule(lr){2-4}  \cmidrule(lr){5-7}\cmidrule(lr){8-8}
 & Hate-Speech &  Sentiment & Toxicity &  Engaging &  Fact-Claiming &  Toxicity &  Stance &   Average   \\  \midrule
 GBERT-base  &  51.7$\pm$4.1 & 59.6$\pm$3.7 &  51.1$\pm$2.5 &   57.8$\pm$1.8 &    70.3$\pm$1.1 &   54.0$\pm$9.2 &   76.0$\pm$2.1 &  60.1 \\
 GBERT-large &  47.6$\pm$4.6 & \textbf{63.0}$\pm$4.8 &  \textbf{54.3}$\pm$3.1 &   \textbf{62.1}$\pm$2.3 &    69.9$\pm$2.0 &   54.6$\pm$3.4 &   \textbf{78.5}$\pm$1.7 &  \textbf{61.4} \\
 GELECTRA-base  &  51.4$\pm$3.4 & \textbf{64.7}$\pm$4.5 &  \textbf{52.0}$\pm$3.1 &   59.4$\pm$1.8 &    \textbf{70.9}$\pm$1.9 &   53.7$\pm$7.1 &   77.1$\pm$1.9 &  \textbf{61.3} \\
 GELECTRA-large &  54.1$\pm$3.6 & \textbf{62.0}$\pm$5.1 &  47.3$\pm$17.3 &   \textbf{61.6}$\pm$2.1 &    68.6$\pm$1.8 &   53.0$\pm$12.1 &   \textbf{78.2}$\pm$1.4 &  60.7 \\ \midrule
 mBERT-base  &  40.4$\pm$5.6 & 41.7$\pm$1.7 &  50.7$\pm$4.3 &   59.2$\pm$2.1 &    70.4$\pm$2.2 &   47.0$\pm$7.4 &   72.0$\pm$2.4 &  54.5 \\
 XLM-R-base  &  46.0$\pm$3.0 & 45.6$\pm$9.0 &  46.8$\pm$17.0 &   59.5$\pm$1.8 &    70.4$\pm$1.7 &   34.2$\pm$29.6 &   74.1$\pm$1.4 &  53.8 \\
 mDeBERTa-base  &  53.1$\pm$4.5 & 63.1$\pm$3.1 &  50.5$\pm$2.6 &   \textbf{60.3}$\pm$2.1 &    \textbf{72.6}$\pm$1.8 &   58.0$\pm$4.9 &   77.8$\pm$1.6 &  \textbf{62.2} \\\midrule
 BERT-base   &  28.0$\pm$6.9 & 41.2$\pm$1.7 &  34.5$\pm$14.1 &   57.7$\pm$3.1 &    66.8$\pm$2.2 &   22.5$\pm$19.8 &   64.2$\pm$ 3.2 &  45.0 \\
 RoBERTa-base   &  41.6$\pm$5.9 & 43.0$\pm$1.2 &  31.9$\pm$19.2 &   57.7$\pm$1.5 &    67.8$\pm$2.8 &   29.7$\pm$5.3 &   66.6$\pm$1.7 &  48.3 \\
 DeBERTa-base   &  52.3$\pm$3.5 & 50.1$\pm$7.7 &  50.8$\pm$2.9 &   59.3$\pm$2.2 &    \textbf{71.7}$\pm$2.5 &   42.0$\pm$18.2 &   \textbf{79.3}$\pm$2.6 &  57.9 \\\midrule
 Sauerkraut-8B  &  \textbf{54.7}$\pm$2.8 & 44.1$\pm$11.4 &  49.9$\pm$10.0 &   56.2$\pm$3.3 &    52.7$\pm$7.0 &   58.7$\pm$4.5 &   55.3$\pm$1.8 &  53.1 \\
 Sauerkraut-70B &  \textbf{56.1}$\pm$1.4 & 43.2$\pm$10.6 &  46.6$\pm$7.4 &   50.7$\pm$3.4 &    59.3$\pm$13.2 &   \textbf{67.9}$\pm$3.2 &   56.9$\pm$7.0 &  54.4 \\\midrule
 Llama-3-8B  &  42.0$\pm$5.8 & 35.1$\pm$23.5 &  27.9$\pm$5.9 &   46.0$\pm$7.0 &    59.7$\pm$4.4 &   51.0$\pm$8.3 &   52.5$\pm$11.7 &  44.9 \\
 Llama-3-70B &  57.2$\pm$3.9 & 38.8$\pm$13.3 &  41.9$\pm$6.9 &   50.7$\pm$2.6 &    62.9$\pm$12.9 &   \textbf{68.7}$\pm$2.8 &   57.4$\pm$5.6 &  53.9 \\
 gpt-3.5-turbo  &  55.3$\pm$3.9 & 56.6$\pm$3.2 &  26.7$\pm$6.7 &   50.3$\pm$1.2 &    44.4$\pm$3.9 &   57.0$\pm$3.3 &   52.8$\pm$3.0 &  49.0 \\
 gpt-4o   &  \textbf{64.7}$\pm$6.2 & 52.5$\pm$4.4 &  \textbf{53.0}$\pm$8.7 &   43.8$\pm$2.7 &    66.5$\pm$14.4 &   \textbf{67.9}$\pm$0.8 &   62.1$\pm$6.5 &  58.7 \\\midrule\midrule
 Average   &  48.2 & 51.8 &  45.8 &   57.6 &    67.5 &   48.2 &   70.9 &  55.7 \\
 \bottomrule
 \end{tabular}
 }
 \caption{Detailed performance on the seven \Lou{} tasks for all the analyzed LMs on the original examples, without reformulations.}
 \label{tab:detailed_baseline}
\end{table*}

\subsection{Label Verification}\label{subsec:app:label-verification}
We list in \autoref{tab:label_verification} manually check examples. We found that gender-fair language does not invalidate any annotated task label. 

\begin{table*}[t]
\centering
    \setlength{\tabcolsep}{3pt}
    \resizebox{0.95\textwidth}{!}{%
\begin{tabular}{lL{19cm}ll}
    \toprule
     \bf Task & \bf Text & \bf Topic & \bf Label\\ \toprule

Stance & Staatlicher Zwang ist falsch. Das ist Sache zwischen Arbeitgeber*in und Arbeitnehmer*in & Welfare & \texttt{against}\\
Stance & Freie Wirtschaft für freie Bürger*innen. Weltweiter Freihandel ohne Schranken ist erstrebenswert. & Foreign Policy & \texttt{favor}\\
Stance & Nicht unter einem*r Präsident*in, welcher die Rechte anderer mit Füssen tritt und Respektlos gegenüber ändern ist. & Foreign Policy & \texttt{against}\\
Stance & Jede anbietende Person soll an seinem Standort selber entscheiden können wie lange geöffnet sein soll & Economy & \texttt{favor}\\
Stance & Das wäre kontraproduktiv. Das Problem, dass ältere Arbeitnehmer*innen keine Stelle mehr finden, würde dadurch verschärft. & Economy & \texttt{against}\\
Stance & Es konnte kein Rückgang bei kriminellen Straftaten festgestellt werden. \texttt{favor} geraten bislang unbescholtene Bürgerne zunehmend unter Generalerneverdacht. Dieser Entwicklung ist Einhalt zu gebieten. & Security & \texttt{against}\\
Stance & Es sollen Anreize geschaffen werden (z.B. via BVG-Beiträge) damit es für Arbeitgeberne attraktiv bleibt, ältere Angestellte im Betrieb zu behalten. Ein Kündigungsschutz setzt falsche Anreize. & Economy & \texttt{against}\\
Stance & Es sollen die gleichen Spielregeln für alle gelten – die Online Anbieter bewegen sich oft noch im Grau-Bereich. Die Angebote sollen aber nicht durch Regulierungen verunmöglicht werden. & Digitisation & \texttt{favor}\\
Stance & Die Schweiz bietet den internationalen Unternehmen anderweitig genug gewichtige Vorteile (politische Stabilität, Einstellung der Mitarbeiterinnen und Mitarbeiter, Infrastruktur...) & Finances & \texttt{against}\\
Stance & Das Rentenalter soll flexibel sein, so kann jeder eigenverantwortlich bestimmen, Arbeitskräfte wie Unternehmen & Welfare & \texttt{favor}\\
Stance & Wichtig ist , dass alle am Markt teilnehmenden Personen gleich lange Spiesse haben. & Digitisation & \texttt{favor}\\\midrule

Fact-Claiming & @USER stimmt. Die Russ*innen hatten wenigstens diese Ossis unter Kontrolle & & \texttt{no fact claimed}\\
Fact-Claiming & Tja, nur weil das bei uns so gehandhabt wird wenn die Wahl zum Staatsoberhaupt nicht passt, heißt das noch lange nicht das rs überall auf der Welt so läuft & & \texttt{fact claimed}\\
Fact-Claiming & Das hoffen die allermeisten meiner amerikanischen Lieblingsmenschen  allerdings nicht. Vote him out. & & \texttt{no fact claimed}\\
Fact-Claiming & Republicans Overseas haben sich echt nicht positiv hervorgetan die letzten Wochen, das ist an Peinlichkeit kaum zu überbieten. Ankündigungen und Lügen. Super Staatsoberhaupt habt ihr da. & & \texttt{no fact claimed}\\
Fact-Claiming & @USER, bin ich Politikere und verdiene jede Menge? Nein. & & \texttt{no fact claimed}\\
Fact-Claiming & @USER und eigentlich, ja eigentlich hätte sie gewonnen, wenn nicht das amerikanische Wahlsystem keine eigentliche Gewinnerin oder eigentlichen Gewinner kennen würde! & & \texttt{fact claimed}\\
Fact-Claiming & Warum ist danach Schluss? F. D. Roosevelt war auch 3 Amtszeiten Staatsoberhaupt. Das Gesetz wird der Orange wohl auch noch einmal ändern & & \texttt{fact claimed}\\
Fact-Claiming & Er ist der aller schlimmste Präsident den Amerika je hatte .... & & \texttt{no fact claimed}\\
Fact-Claiming & Hackt nicht nimmer auf den Fussball rum. Bei allen Sportarten ist wieder Publikum erlaubt. Hygienekonzept vorausgesetzt. & & \texttt{fact claimed}\\
Fact-Claiming & Impfkritische Personen haha,wir haben einfach nur angst um unsere Gesundheit. Ich will nicht durch ne Impfung krank werden in 2 oder 3 Jahren. & & \texttt{no fact claimed}\\\midrule

Engaging & Genau wegen solchen Eingeladenen in dieser Sendung schaue ich mir @MEDIUM nicht mehr an, eine Zumutung ist das. & & \texttt{engaging}\\
Engaging & George Weinberg war so ignorant und abstoßend, dass er nen gutes Sinnbild für Trumps Lager abgegeben hat. Die Republikaner*innen sind seit Bush in einer stetigen menschlichen Abwärtsspirale. & & \texttt{not engaging}\\
Engaging & Wie ist das mit der Zwangsabgabe- Rundfunkbeitrag? Zahlt die doch direkt an die Schauspielerne und Künstlerne aus! Dann sind sie sinnvoller eingesetzt! & & \texttt{engaging}\\
Engaging & Regierende mit Verstand ?...und das in Zusammenhang mit dem Namen Trump ? & & \texttt{not engaging}\\
Engaging & Was heisst dort lassen sollen? Der ist nur nach Deutschland gekommen weil es die Russ*innen so erlaubt haben. Deutschland hatte da garnicht's zu entscheiden. & & \texttt{engaging}\\
Engaging & Fussballclubs bestehen nicht nur aus ein paar Kickerinnen oder Kickern und Trainerinnen oder Trainern. Was Sie fordern würde Tausende Angestellte treffen. & & \texttt{not engaging}\\
Engaging & @USER Auffallend viele Republikanerne können mit enserm Präsidente nichts (mehr) anfangen. & & \texttt{not engaging}\\\midrule

Hate-Speech & @ZDFheute Mit dem Bericht aus Israel, wo gegen Israel gehetzt wird, zeigt sich, wer hier antisemitisch ist. Ist jetzt Aktivistin ein neuer Beruf für schreiende, ungezogene, rechthaberische Mädchen? & & \texttt{no hate-speech}\\
Hate-Speech & @TsiJulian @janboehm Promovierte Journalistinnen und Journalisten. Na und? & & \texttt{no hate-speech}\\
Hate-Speech & @MichaelDhr1 @kristen\_peter @Xrxtn @Alice\_Weidel @UweJunge Steht ebend nicht drin, meine Göttin oder mein Gott. Die Binde muss nur zu erkennen sein. & & \texttt{no hate-speech}\\
Hate-Speech &	@ubskm\_de @AfD\_RBK Sehr gut. Und bitte noch die diversen Schlupflöcher weitgehend zukitten, damit kein linksgrün verstrahlte "Sachverständigere" de Tätere am Ende doch noch zum Opfer macht.& &	\texttt{hate-speech}\\
Hate-Speech & @AliCologne Die Demo-Teilnehmerne verhöhnen die Corona-Opfer ebenso wie \#Querdenkerne und \#Coronaschwurblerne. Es ist ein Schlag ins Gesicht der Pflegekräfte und Arzterne in den Kliniken. & & \texttt{no hate-speech}\\
Hate-Speech &	@Hoellenaufsicht @eduardt\_s @Alice\_Weidel @UweJunge Benennt diese AFD Leute als das was sie sind Faschisterne, nicht rechts radikal oder sonstiges rumgeiere. Diese Leute sind brandgefährlich Polizei& &	\texttt{hate-speech}\\
Hate-Speech & Ab 1. Juni hebt \#Israel viele Vorteile für \#greenpass Besitzerinnen und Besitzer auf und gestattet weitestgehende \#Lockerungen auch für Nicht- \#Geimpfte . \#Corona https://t.co/wlZ5tmt3HJ & & \texttt{no hate-speech}\\
Hate-Speech &	@Svenfpunkt @LillyBlaudszun @janboehm 76 Wählerinnen und Wähler sind jetzt nicht die Welt, aber das Dorf sollte man bei Gelegenheit mal entseuchen.& &	\texttt{hate-speech}\\
Hate-Speech &	@\_FriedrichMerz Die vielen Rechtschreib- und Orthographie Fehler in den Anti-Merz-Tweets beweisen das geistige Kleinrentnerinnen- und Kleinrentnertum der linksgrünen Absenderinnen und Absender. & &	\texttt{hate-speech}\\\midrule

Toxicity & Ich hoffe ihre Gäste werden Recht behalten und Jo Biden wird  Präsident. Noch mal 4 Jahre Trump und der bringt die ganze Welt durcheinander & & \texttt{not toxic}\\
Toxicity &	kein Wunder bei diesen Haufen lügenden und betrügenden Personen die sich Titeln kaufen und ein auf schlau machen !!!	& &\texttt{toxic}\\
Toxicity & @Anna48338034 @mista98berk Frag mal Arye Sharuz Shalicar, er ist hier auf Twitter und Sprecher der israelischen Armee, in Deutschland geboren, er kann das bestätigen & & \texttt{not toxic}\\
Toxicity &	Tja, nur weil das bei uns so gehandhabt wird wenn die Wahl zur Ministerpräsidentin oder Ministerpräsidenten nicht passt, heißt das noch lange nicht das rs überall auf der Welt so läuft	& &\texttt{toxic}\\
Toxicity & @USER Die nächste Eiszeit kommt. Derzeit leben wir in einer Zwischenwarmzeit. Durchaus seriöse forschende Personen halten ein schnelles Abreisen des Golfstroms für sehr wahrscheinlich. & & \texttt{not toxic}\\
Toxicity &	https://t.co/cs5KoavBp8  NRW Lusche ihr seid scheiße nein du bist es! Eine Bekanntschaft aller Schwulen Spahnferkels. thx& &	\texttt{toxic}\\
Toxicity & Sein Wort in Gött*in Gehörgang  & & \texttt{not toxic}\\
Toxicity &	@willensbrecher Willensbrecher du bist ein kranker und ekelhafter Spinner oder eine kranke und ekelhafte Spinnerin verzieh dich aus Twitter oder ich melde dein Tweet den Behörden du Nazi& &	\texttt{toxic}\\
Toxicity & Da bin ich ja mal gespannt. Denn das, was ich von den Lernenden und jungen Leuten vor allem mitbekomme ist, dass sie sich wie die Versuchskaninchen für das Virus vorkommen. & & \texttt{not toxic}\\
Toxicity & 	@MEDIUM Warum denn mit fiebern? "Die Amerikanerinnen und Amerikaner werden schon das Richtige machen!.... nachdem sie alles andere Ausprobiert haben."	& & \texttt{toxic}\\
Toxicity &	Die USA würden sich viel Stress ersparen, wenn sie einfach das nächste Staatsoberhaupt vom @MEDIUM und @MEDIUM wählen lassen würden.	& & \texttt{toxic}\\
Toxicity &	@Svenfpunkt @LillyBlaudszun @janboehm 76 Wählende sind jetzt nicht die Welt, aber das Dorf sollte man bei Gelegenheit mal entseuchen.	& & \texttt{toxic}\\
Toxicity &	@TiloJung Als regierungssprechende Person muss man dumm sein. BÖSE und DUMM. \#niewiederCDU \#fckcdu	& & \texttt{toxic}\\\midrule

Sentiment &@BastardHegels @iknrr @ainyrockstar Alter der Syrer saß da ganz friedlich im Bus wie jeder andere auch und der Nazi  Spast kommt an und attackiert ihn wtf ist falsch bei dir  & &  \texttt{negative}\\
Sentiment &@sschyvonne @AndySpirig @Karl\_Lauterbach Herr, lass Hirn über Frau*Herr Richter*in regnen! & & \texttt{negative}\\
Sentiment &@MarkusWerner18 @c\_lindner Der Georg Thile hat bekommen , was er bekommen soll, wer GEZ nicht bezahlt bricht das Gesetz, also eine Perso , die Verbrechen begeht. & &  \texttt{negative}\\
Sentiment &@Svenfpunkt @LillyBlaudszun @janboehm 76 Wählerne sind jetzt nicht die Welt, aber das Dorf sollte mensch bei Gelegenheit mal entseuchen. & &  \texttt{negative}\\

    \bottomrule
    \end{tabular}
    }
    \caption{Overview of the label verification. We randomly chose these examples from the original and the reformulated examples. We manually checked them and found that the annotated task labels do not change.}
    \label{tab:label_verification}
\end{table*}

\subsection{Detailed Reformulation Distribution}\label{subsec:app:detail-analysis}

\begin{figure*}[t]
 \centering
 \includegraphics[width=1\textwidth]{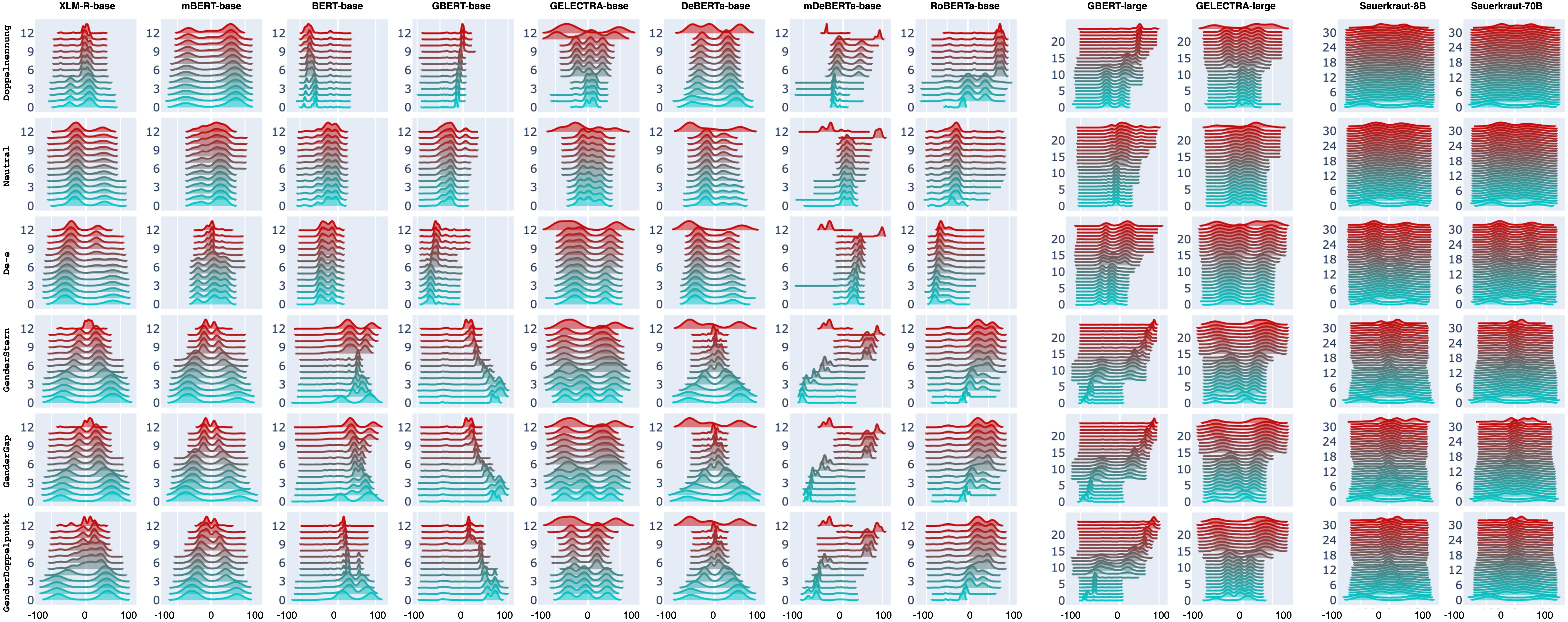}
 \caption{Projection (1D using T-SNE) of the vector difference between the average embeddings of the reformulated examples and the original ones for all six strategies and 13 layers (x-axis) of GBERT-base, including the embedding layer (0).}
 \label{fig:1d_embeddings_all}
\end{figure*}

\autoref{fig:1d_embeddings_all} shows the distribution of the reformulation representation $r$ for every model, reformulation strategy and model layer.
Similar patterns, as previously discussed, can be observed: strategies are more distinguishable for lower layers, and noteworthy differences between \GenderDoppelpunkt{} and \GenderStern{} and \GenderGap{}.
Further, LMs with a higher performance level (like German LMs) tend to show more variation among the layers, hinting at their better semantic understanding of the German language.  

\end{document}